\documentclass[10pt,twocolumn,letterpaper]{article}

\usepackage{3dv}
\usepackage{times}
\usepackage{epsfig}
\usepackage{subfigure}
\usepackage{graphicx}
\usepackage{amsmath}
\usepackage{amssymb}
\usepackage{algorithm}
\usepackage{algpseudocode}
\usepackage[utf8]{inputenc}
\usepackage[T1]{fontenc}

\usepackage[pagebackref=true,breaklinks=true,colorlinks,bookmarks=false]{hyperref}

\threedvfinalcopy % *** Uncomment this line for the final submission

\def\threedvPaperID{174} % *** Enter the 3DV Paper ID here

\ifthreedvfinal\fi

\DeclareMathOperator{\mean}{mean}

\algnewcommand\INPUT{\item[\textbf{Input:}]}%
\algnewcommand\OUTPUT{\item[\textbf{Output:}]}%

\begin{document}

%%%%%%%%% TITLE
\title{Novel Single View Constraints for Manhattan 3D Line Reconstruction}
\author{
Siddhant Ranade \quad Srikumar Ramalingam \\
University of Utah, USA
\\{\tt\small \{sidra,srikumar\}@cs.utah.edu
}
}

%\author{Siddhant Ranade\\
%University of Utah\\
%{\tt\small sidra@cs.utah.edu}
%\and
%Srikumar Ramalingam\\
%University of Utah\\
%{\tt\small srikumar@cs.utah.edu}
%}

\maketitle
% \thispagestyle{empty}

%%%%%%%%% ABSTRACT
\begin{abstract}
This paper proposes a novel and exact method to reconstruct line-based 3D structure from a single image using Manhattan world assumption. This problem is a distinctly unsolved problem because there can be multiple 3D reconstructions from a single image. Thus, we are often forced to look for priors like Manhattan world assumption and common scene structures. In addition to the standard orthogonality, perspective projection, and parallelism constraints, we investigate a few novel constraints based on the physical realizability of the 3D scene structure. We treat the line segments in the image to be part of a graph similar to \emph{\bf straws and connectors game}, where the goal is to back-project the line segments in 3D space and while ensuring that some of these 3D line segments connect with each other (i.e., truly intersect in 3D space) to form the 3D structure. We consider three sets of novel constraints while solving the reconstruction: (1) constraints on a series of Manhattan line intersections that form cycles, but are not all physically realizable, (2) constraints on true and false intersections in the case of nearby lines lying on the same Manhattan plane, and (3) constraints from the intersections on boundary and non-boundary line segments. The reconstruction is achieved using mixed integer linear programming (MILP), and we show compelling results on real images. Along with this paper, we  will release a challenging Single View Line Reconstruction dataset with ground truth 3D line models for research purposes.
\end{abstract}

%%%%%%%%% BODY TEXT
\section{Introduction}
As Sugihara~\cite{sugihara1986} points out:

\begin{quote}
\emph{Human beings invented a noble class of pictures called ``line drawings'' as a means of representing three-dimensional shape of objects.}
\end{quote}

\begin{figure}[!t]
\centering
\psfig{figure=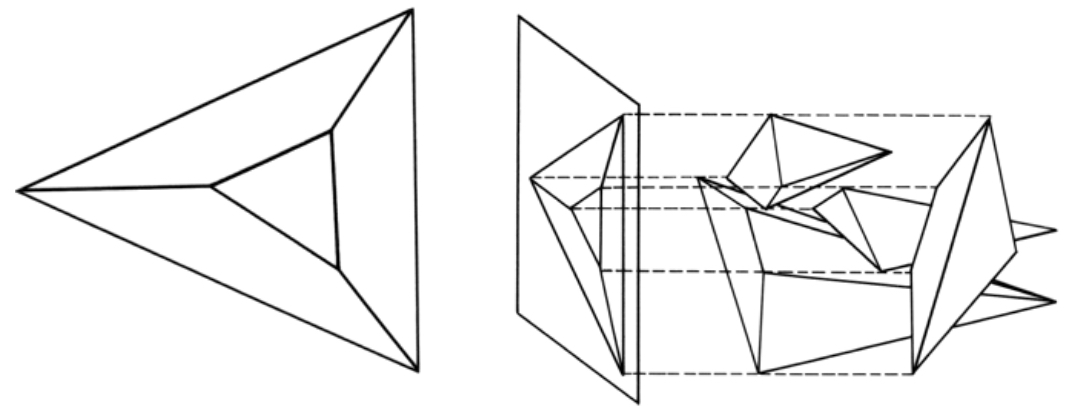,width=1.0\columnwidth}
\caption{\it Left: Line drawing of a physically realizable truncated tetrahedron with truncation at one of the vertices. Right: Several polyhedrons placed at different distances in 3D space can project to form the image of a truncated tetrahedron. The figure is adapted from ~\cite{sugihara1986}}
\label{fg:t_tetrahedron}
\end{figure}

The research in line drawings spans several domains (geology, engineering drawing, human communication, computed aided systems, art, etc.). Line drawings can be seen as a visualization tool that shows human interpretable diagrams with reduced dimensions compared to original image or other sensor data. 

Consider an example in Fig.~\ref{fg:t_tetrahedron}. On the left, we show the projection of the truncated tetrahedron. On the right, we show one possible 3D model where a set of polyhedrons in 3D space project to form an image of truncated tetrahedron. While the human brain can interpret the geometry of the underlying 3D object without much effort (on most line drawings), how do we make the computer identify the most appropriate interpretation of the line drawings? This problem is one of the classic problems in 3D reconstruction, ever since Robert built a system to identify the 3D prototype whose projection matches with the line drawings in 2D~\cite{roberts1965}. In addition to computer vision, several other domains such as constraint satisfaction~\cite{Cooper2008} focused on developing novel algorithms for line drawing interpretation, and used this problem as their test case. While classical algorithms heavily relied on identification of novel and intricate constraints that can handle several pathological cases, not much effort was made to apply these methods on real data. Many of the modern methods have completely ignored the classical results, and they are focused on developing simple "black-box" algorithms that seem to produce good pixel-wise results on large datasets, but not 3D models with precise boundaries. This makes us wonder as to whether we can borrow some useful ideas from these classical works for reconstructing from real world data.

The term ``3D Reconstruction'' is defined in many ways. Very often, it refers to an algorithm that generates some geometrical entities/coordinates, such as the extraction of sparse/dense point cloud or plane-based 3D models. However, limited work has been done in the context of identifying all the underlying constraints in the geometrical model. We argue that an accurate 3D model not only generates the 3D coordinates of most points in the object, but also develops CAD-style or polyhedral 3D models to obtain the associated line segments, their orientations, angles and intersections with other line segments, as well as their physical locations.  

\begin{figure}[t]
\centering
\psfig{figure=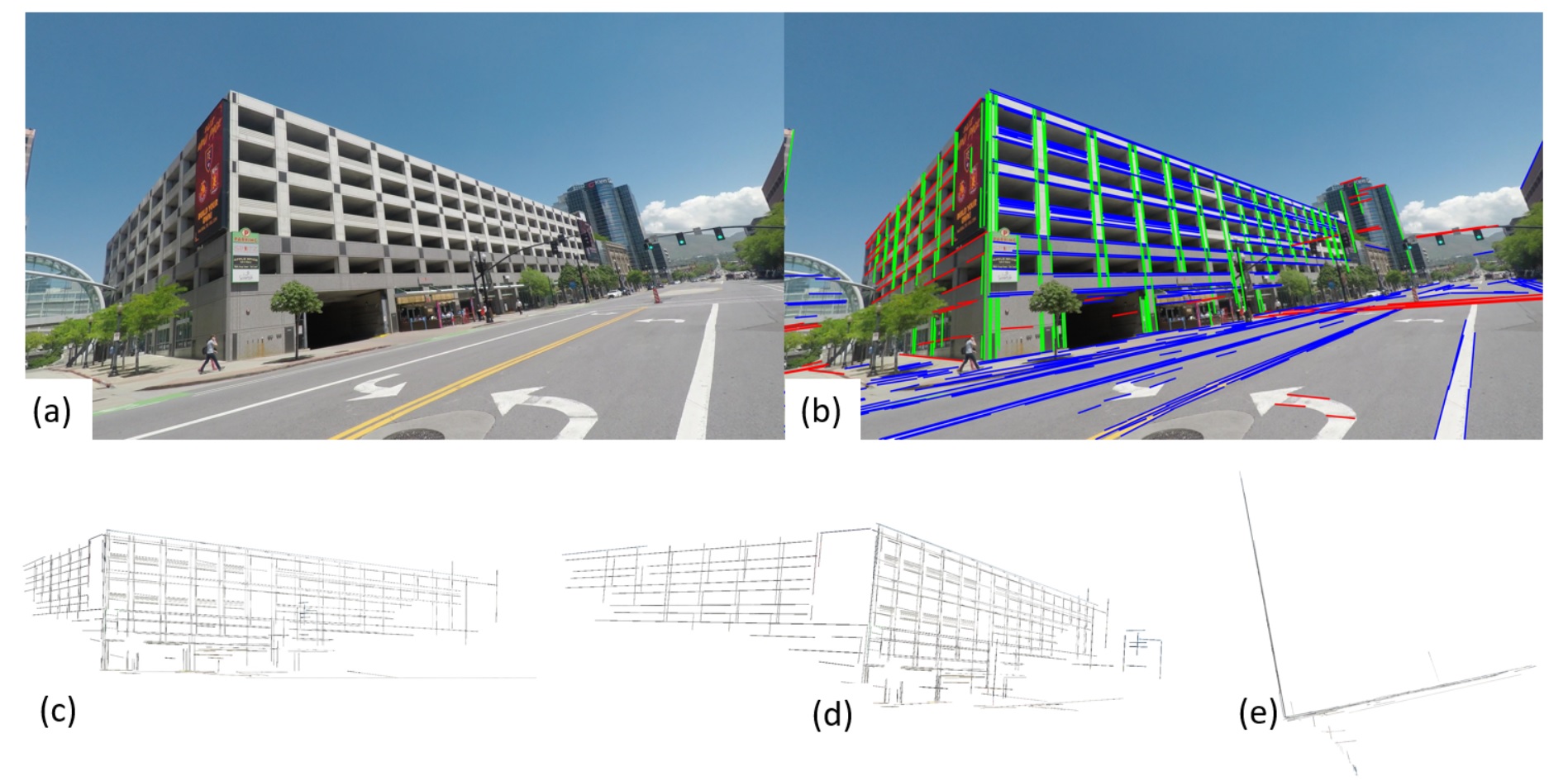,width=1.0\columnwidth}
\caption{\it (a) We show an image of an outdoor scene. The red, green and blue lines denote the line segments that are directed along $x$, $y$ and $z$ directions in the 3D space. (b) While several 2D line segments intersect in the image space, a subset of these intersections also occur in 3D. We propose novel constraints to identify these "true" intersections in 3D to reconstruct the object in 3D space as shown in different perspective views (c),(d) and (e).}
\label{fg.frontPageIntroFigure}
\end{figure}

%Strategy 2:
%Loop constraints are not explored in 3D reconstruction. Bring in some ideas. from Martin Cooper. Check Sugihara's tutorial for some motivation. Talk about some classical papers talking about loop constraints and not using them. This is the first paper that exploits that in real world images.

%Strengths: 
%* MILP formulation.

Consider Fig.~\ref{fg.frontPageIntroFigure}. We show the original image and the Manhattan line segments detected on the image. Using the line reconstruction algorithm proposed in this paper, we show the results from three different viewpoints in (c),(d), and (e). Our approach builds on top of the line-lifting framework using Manhattan line segments and identifying the connectivity information~\cite{Ramalingam2013_lifting}. The basic idea in \cite{Ramalingam2013_lifting} is to formulate the 3D reconstruction problem as a Boolean optimization problem where the goal is to identify the correct intersections between nearby line segments in 3D space. In this paper, we propose novel constraints based on the physical realizability of the scene to improve their results. In contrast to the work of \cite{Ramalingam2013_lifting}, which uses $L1$-norm minimization to relax the original $L0-$norm problem, we formulate the reconstruction problem using mixed-integer linear programming (MILP) without any relaxation. While identifying the true connectivity, we formulate the problem on a line graph with line segments as vertices and their potential connections as edges. This paper explores three different constraints in the context of single-view reconstruction for Manhattan worlds. In the first class of constraints, we consider sub-graphs that are cycles and propose novel constraints that rule out certain intersections. In the second class, when a line segment belongs to one of Manhattan planes (i.e., planes oriented in the Manhattan directions), we enforce constraints on certain intersections based on planarity constraints. The third family of constraints enforces boundary constraints: we identify the boundary and non-boundary line segments and prohibit certain types of intersections. These constraints are shown to produce improvement over the existing methods. To the best of our knowledge, no prior work has used such constraints.

\subsection{Contributions}

The main contributions of this paper are as follows:
\begin{itemize}\setlength\itemsep{0em}
\item We propose cycle graph constraints for enabling single view 3D reconstruction. 
\item We propose planarity constraints based on nearby line segments lying on planes oriented along Manhattan directions.
\item We propose boundary constraints on the intersections, based on whether certain line segments are boundary or non-boundary ones. 
\item We formulate the problem as a mixed-integer linear programming (MILP) that incorporates the hard geometrical constraints proposed in this paper. 
\item We show compelling real world reconstruction results and show both qualitative and quantitative improvement over existing methods. 
\end{itemize}

\section{Related Work}
Though single view reconstruction (\textsc{SVR}) is a classical problem in computer vision~\cite{kanade1980,malik1987,roberts1965,sugihara1986,waltz1972}, it is still an actively researched topic to this date. We classify the existing methods into two classes: constraint-driven and data-driven methods.
\smallskip

\noindent
{\bf Constraint-driven classical results.}
Most classical methods were constraint-driven, and in particular, utilized techniques to classify line segments into convex, concave, and occluding labels. The underlying assumptions are that line segments are obtained from the intersection of two planes and the angle between the planes determines the label. Once the line labeling problem is solved using constraint satisfaction algorithms such as backtracking, it is possible to lift the line segments in 3D space~\cite{sugihara1986}. The theory for handling non-planar surfaces and analysis with intersections and junctions from curves was developed a few decades ago~\cite{malik1987}. While the theory has been established for synthetic line drawings, such techniques have never been tested on real world images~\cite{sugihara1986}. This is because many of the assumptions used in classical works do not hold true on real images. First, there are lots of spurious and missing line segments in real images. The connectivity between line segments is not available and there is no straightforward way to get this. In fact, this paper primarily focuses on generating the connectivity between 3D line segments, and this information is assumed to be already available in all the classical approaches. Thus it is not all that surprising to see that a minimal user-interaction to manually provide the connectivity and constraints can help us generate accurate 3D models from a single image~\cite{criminisi2000,sturm99}.
\smallskip

\noindent
{\bf Data-driven methods.}
In contrast to classical results that employ geometrical constraints, recent learning based methods use hundreds of training images to get a mapping between semantic classes and geometry of the scene, i.e., the pixel-wise class labels can provide depth cues. For example, Hoiem refers to this as the geometric layout estimation which looks at the problem as a semantic segmentation algorithm to classify the pixels in the image pixels into sky, buildings and ground~\cite{hoiem2005,hoiem2007}. This classification is already sufficient to generate popup models and enable stereoscopic content generation. Saxena~\etal~\cite{saxena2008} developed an algorithm for depth estimation from a single image using collinearity and coplanarity assumptions. Clutter makes the problem of indoor reconstruction particularly challenging using single views. Hedau~\etal~\cite{hedau2009} used a cuboid approximation and evaluates different hypotheses to identify the best cuboid using structured learning methods. Several novel ideas such as the computation of orientation maps~\cite{lee2010}, inferring geometry from human activities~\cite{fouhey2012} and even physics-driven stability and mechanical constraints~\cite{gupta2010} have been utilized for single view reconstruction. Schwing~\etal~\cite{schwing2012a} used efficient inference machinery to show improvement in the indoor layout estimation algorithm. With the recent resurgence of deep learning methods, an end-to-end deep neural network (DNN) have been proposed to obtain single view 3D reconstruction~\cite{Liu2018}. While recent learning based algorithms are promising, several decades of constraint-based techniques should not be excluded, as they do provide strong tools for obtaining accurate 3D models, that are not easy to obtain using purely learning based techniques.
\smallskip

\noindent
{\bf Recent constraint-based SVR methods.}
One of the most popular constraints, not explored much in classical results~\cite{Perkins1971}, is the so-called Manhattan world prior~\cite{coughlan99}. Most man-made scenes, both indoor and outdoor, satisfy the Manhattan world assumption~\cite{coughlan99}. Delage~\etal~\cite{delage05} used this assumption in an MRF framework to reconstruct indoor scenes. Constraints based on Manhattan grammar have been used for modeling buildings from aerial photos~\cite{vanegas10}. Our work is related to the geometrical constraints used in ~\cite{lee2009,flint2011}. Structures like rectangles were detected in~\cite{han2009,micusik2008}, and they can be ordered according to their depth~\cite{yu2008}. The single view 3D reconstruction is formulated as a model fitting problem and vertical walls and ground plane are 
estimated~\cite{Barinova2008}. Template or example-based approaches have been used to reconstruct 3D scenes from line drawings~\cite{cole2012}. Elasticity constraints have been explored in single view 3D reconstruction of non-planar surfaces~\cite{Malti2013}. Kushal and Seitz developed a single view 3D reconstruction for piece-wise swept surfaces~\cite{Kushal2013}. Recently, it was shown that junction features can be extracted from real images using an efficient voting scheme~\cite{Ramalingam2013CVPR}. Junctions are points where two more or more lines intersect, and based on the number of lines and the angles between them, they can be generally classified as $L$, $Y$, and $X$. This work uses junctions for designing penalty terms in a linear programming (LP) formulation.
\smallskip

\noindent
{\bf Constraint-based multi-view methods.}
We review a few, definitely not all, multi-view reconstruction methods that have exploited interesting constraints in the context of 3D reconstruction using line segments. Jain~\etal~\cite{jain2010} used connectivity constraints for reconstructing lines from multiple images. Sinha \etal~\cite{Sinha2008Interactive3A} used Manhattan priors for obtaining plane-based 3D reconstruction from multiple images. Line reconstruction using multiple images and orientation constraints has been done in ~\cite{Schindler2006LineBasedSF}. 3D reconstruction from point-clouds is formulated as a binary labeling problem to generate accurate 3D models~\cite{Nan2017}.

\section{Method}
% !TEX root = ../paper.tex
%
Given an image of a Manhattan scene, our goal is to reconstruct a line model of the underlying 3D scene. We begin by using a line segment detector \cite{MCMLSD} to detect line segments in the image. In 2D, each line segment $l_i$ can be represented by its end points $p_{i1}$ and $p_{i2}$. We assume that the scene primarily has Manhattan lines -- lines that are oriented along 3 orthogonal directions, henceforth known as the $x,y,z$ directions of the world space. If this assumption holds, we can estimate the vanishing points of these three directions in the image, and hence compute a rotation matrix $R$ that relates the camera coordinates to the world coordinates, provided the intrinsic camera matrix $K$ is known. Using the vanishing points, we can also label each line $l_i$ with its ``true direction'' label $\mathcal{D}_i \in \{x,y,z\}$. Note that this true direction is the actual direction of the underlying 3D line in world space.
In cases where a line coincides with the line joining two vanishing points, the estimated ``true direction'' may be incorrect.

We use little $p_{ia}$ to denote endpoint $a$ of line $l_i$ in 2D, represented in 2D homogeneous coordinates. The same point, in 3D world space is denoted by capital $P_{ia}$ (not homogeneous coordinates). For convenience, we say that the origins of the world and camera spaces coincide at the camera center.
Thus 2D and 3D points are related by
\begin{align}
    P_{ia} &= \lambda_{ia} R^{-1} K^{-1} p_{ia} \\
    &= \lambda_{ia} d_{ia},
\end{align}
where the vector $d_{ia} = R^{-1} K^{-1} p_{ia}$ is the direction of a ray from the origin to $P_{ia}$ in world coordinates, and the scalar $\lambda_{ia}$ is the distance the ray has to travel from the origin to reach $P_{ia}$.

\subsection{Line Graph}

\begin{figure}[ht]
    \centering
    \includegraphics[width=\columnwidth]{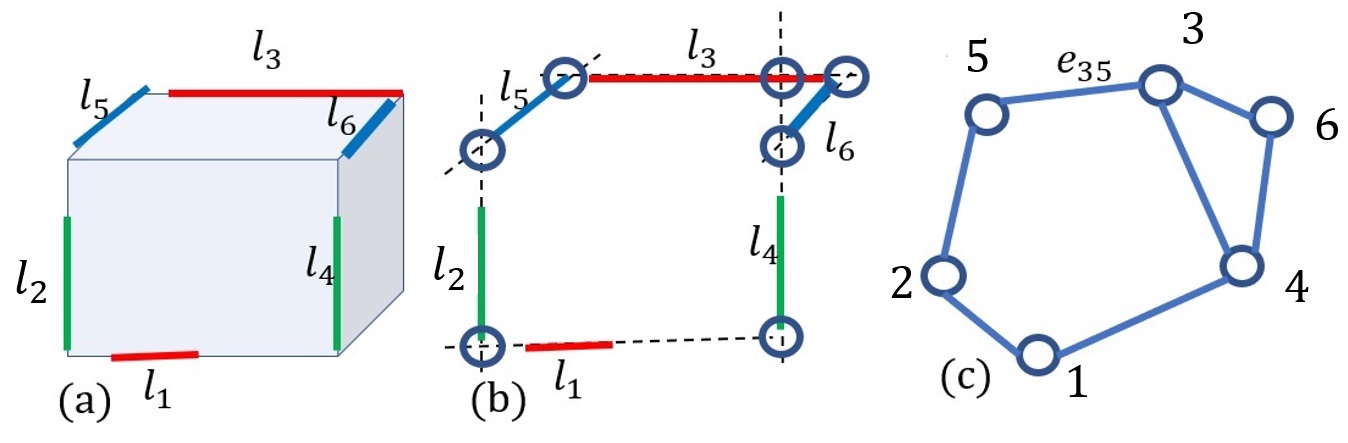}
    \caption{(a) Detected line segments at the boundaries of a cuboid. (b) Extend the line segments to form intersections. (c) A simple line graph where line segments become vertices and intersections between line segments become edges.}
    \label{fig:lineGraphIdea}
\end{figure}

Our approach is that all pairs of line segments which intersect in 2D and which have different true-direction labels are candidates for being an intersecting pair of lines in 3D. We can thus think of a ``line graph'' of the image, where the set of lines $l_i,l_j,\ldots$ correspond to the vertices $\mathcal{V} = \{i,j,\ldots\}$ of the graph, and the set of intersections corresponds to the set of edges $\mathcal{E}$ of the graph as shown in Fig.~\ref{fig:lineGraphIdea}.

\subsection{Reconstruction as Optimization}

\begin{figure}[ht]
    \centering
    \includegraphics[width=0.8\columnwidth]{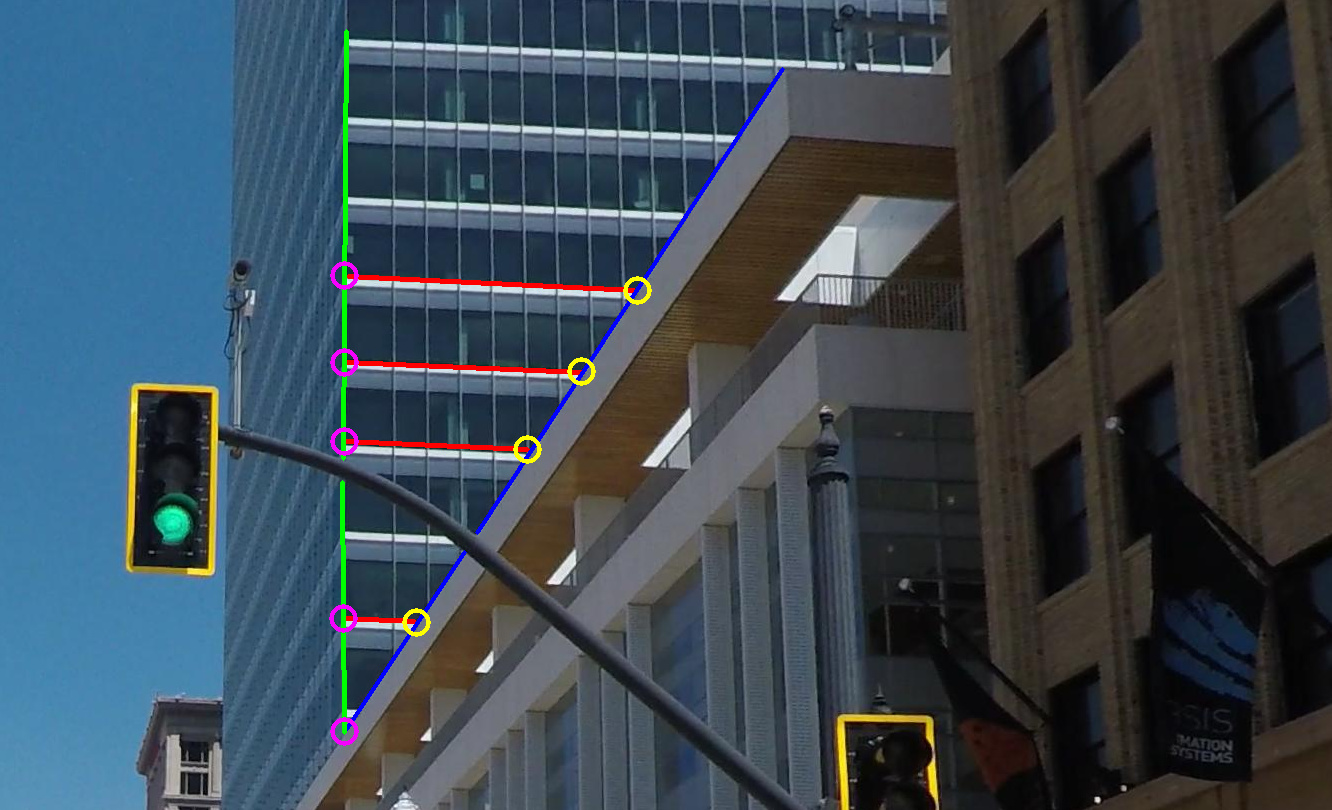}
    \caption{Real and fake intersections (circled with magenta and yellow, respectively). The blue $z$-line does not intersect with the red $x$-lines.}
    \label{fig:RealFake}
\end{figure}

Every edge of the line graph corresponds to either a real intersection in the 3D scene (i.e.\ the corresponding 3D lines actually intersect), or not (See Fig.~\ref{fig:RealFake}). Consequently, we associate with each edge $(i,j) \in \mathcal{E}$ a boolean variable $b_{ij}$, where a value of $1$ indicates a real intersection. Since the ground truth values of the Boolean variables are unknown, we pose this as an optimization problem.
We argue that we want as many edges as possible to correspond to actual intersections, and we maximize the (weighted) number of real intersections in the scene. The optimization is over $\lambda_{ia}$ and $b_{ij}$ variables; the vectors $d_{ia}$, and the matrices $R$ and $K$ are known constants. This optimization is subject to Manhattan constraints. Since all of our constraints and the objective are linear, the optimization can be expressed as a mixed integer linear program (MILP)

\begin{align}
    & \max_{\lambda_{ia}, b_{ij}} \sum_{(i,j) \in \mathcal{E}} w_{ij} b_{ij} \label{eqn:LP-obj}\\
    \text{s.t.} \quad
    & \left\lvert \lambda_{ia} d_{ia\alpha} - \lambda_{jb} d_{jb\alpha} \right\rvert \leq L(1-b_{ij}), \quad a,b \in \{1,2\}, \nonumber \\
    & \alpha = \{x,y,z\} \setminus \{\mathcal{D}_i, \mathcal{D}_j\}, (i,j) \in \mathcal{E} \label{eqn:MILP-cons-1}\\
    & \left\lvert \lambda_{i1} d_{i1\beta} - \lambda_{i2} d_{i2\beta} \right\rvert \leq 0, \nonumber \\
    & \quad \beta \in \{x,y,z\} \setminus \{\mathcal{D}_i\}, i \in \mathcal{V} \label{eqn:MILP-cons-2}\\
    & \lambda_{ia} \geq 1,  \quad a \in \{1,2\}, i \in \mathcal{V}, \label{eqn:MILP-cons-3}
\end{align}
where $w_{ij}$ is the weight associated with every intersection, $L$ is a large constant, and $d_{ia\alpha}$ represents the $\alpha$-component of the direction vector of the end point $a$ of line $l_i$. The weights can be derived from Manhattan junctions, which are points in the images where two or more lines intersect. In the work of ~\cite{Ramalingam2013_lifting}, a voting based junction detector~\cite{Ramalingam2013CVPR} was employed. In this paper, we use a simple algorithm that just looks for intersection of line segments to classify the junctions into $L$, $Y$, and so on. We use the same weights that were used in ~\cite{Ramalingam2013_lifting} for intersections associated with different junctions.

The first constraint says that if a candidate pair of lines is intersecting ($b_{ij}=1$), then the component of the vector joining their end points along the third, mutually perpendicular direction is $0$. $L$ is a large constant. 
That is, if $\mathcal{D}_i = x$, and $\mathcal{D}_j = y$, then the $z$-component of the vector $(\lambda_{ia} d_{ia} - \lambda_{jb} d_{jb})$, i.e., $(\lambda_{ia} d_{iaz} - \lambda_{jb} d_{jbz})$ is $0$.

The second constraint says that the projection of a line segment along directions other than its true direction is $0$, and the third constraint forces all the points to be in front of the camera, beyond a certain distance (as opposed to being behind it).
Note that this constraint is essential because without it, $\lambda_{ia} = 0$ trivially maximizes the objective function to $\sum_{(i,j) \in \mathcal{E}} w_{ij}$, it's maximum possible value.

Multiple disconnected components in the line graph can not be reconstructed in the same scale, and thus we only show the reconstruction of the largest connected component of the line graph, up to a scale factor.

%This basic formulation can not be expected to work very well, because some of the intersections in the image do not correspond to actual intersections.

Though not all intersections in the image correspond to actual intersections in the scene, it is possible to come up with a set of feasible intersections by enforcing a set of constraints which are described in what follows.

\subsection{Cycle Graph Constraints}
\begin{figure}[t]
    \includegraphics[width=\columnwidth]{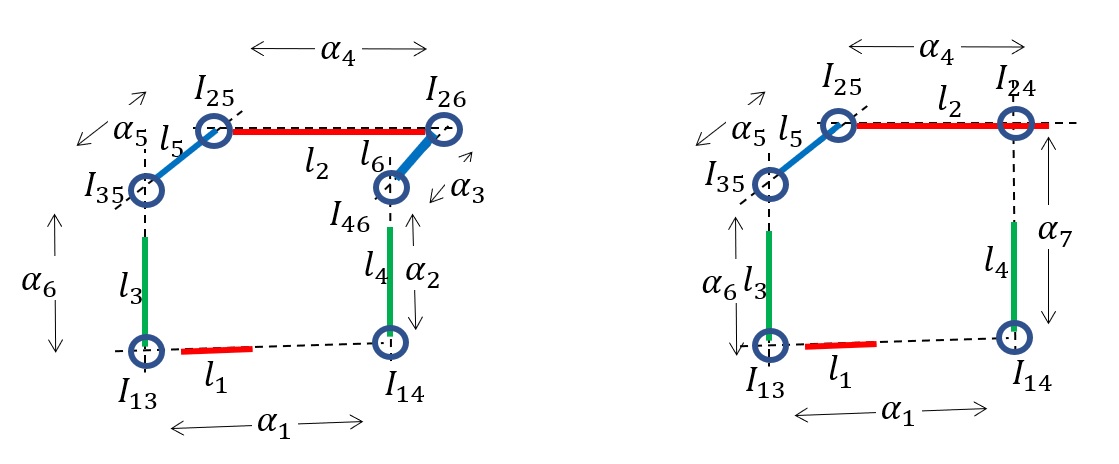}
    \caption{(a) A series of line intersections forming a cycle graph where all the intersections may happen in a physically realizable structure. (b) A cycle graph where all the intersections can not all hold true.}
    \label{fig:cycleGraphConstraint}
\end{figure}

Consider Fig.~\ref{fig:cycleGraphConstraint}. The red, green, and blue colored lines denote the directions along $x$, $y$, and $z$ axes in the world coordinate frame. In (a) we observe a series of intersections that form a cycle graph. Let us assume that the lengths of line segments are given by $\alpha$'s. Since the structure is only known up to a scale, we assume that $\alpha_i \ge 1$. Let us assume that the 3D point associated with a reference point, say $I_{13}$ is given by $(x_{13},y_{13},z_{13})^T$. By traversing the cycle and returning to the point $I_{13}$, we can write

\begin{eqnarray}
\begin{pmatrix} x_{13} \\ y_{13} \\ z_{13} \end{pmatrix} & = &
\begin{pmatrix} x_{13} \\ y_{13} \\ z_{13} \end{pmatrix} +
\alpha_1 \begin{pmatrix} 1 \\ 0 \\ 0 \end{pmatrix} +
\alpha_2 \begin{pmatrix} 0 \\ 1 \\ 0 \end{pmatrix} +
\alpha_3 \begin{pmatrix} 0 \\ 0 \\ 1 \end{pmatrix} \nonumber \\
&& -\alpha_4 \begin{pmatrix} 1 \\ 0 \\ 0 \end{pmatrix}
-\alpha_5 \begin{pmatrix} 0 \\ 0 \\ 1 \end{pmatrix}
-\alpha_6 \begin{pmatrix} 0 \\ 1 \\ 0 \end{pmatrix}
\end{eqnarray}
Since we start and end at the same point, we have the following equations from the three coordinates of the point:
\begin{equation}
\{
\alpha_1 - \alpha_4 = 0,~
\alpha_2 - \alpha_6 = 0,~
\alpha_3 - \alpha_5 = 0 \},
\end{equation}
where all $\alpha$'s are positive. It is evident that several solutions satisfy these equations. However, in Fig.~\ref{fig:cycleGraphConstraint}(b), we have only one line segment along the $z$ direction. In order to start at point $I_{13}$ and end at the same point traversing along the cycle graph, we have following conditions:
\begin{equation}
\{
\alpha_1 - \alpha_4 = 0,~
\alpha_7 - \alpha_6 = 0,~
\alpha_5 = 0 \}
\end{equation}
The above set of equations is infeasible because $\alpha_i \ge 1$. From the above example, it is also evident that if the cycle graph consists of only one line segment along one of the three Manhattan directions, then all the intersections can not hold true. 

\begin{figure}[ht]
    \centering
    \includegraphics[width=0.8\columnwidth]{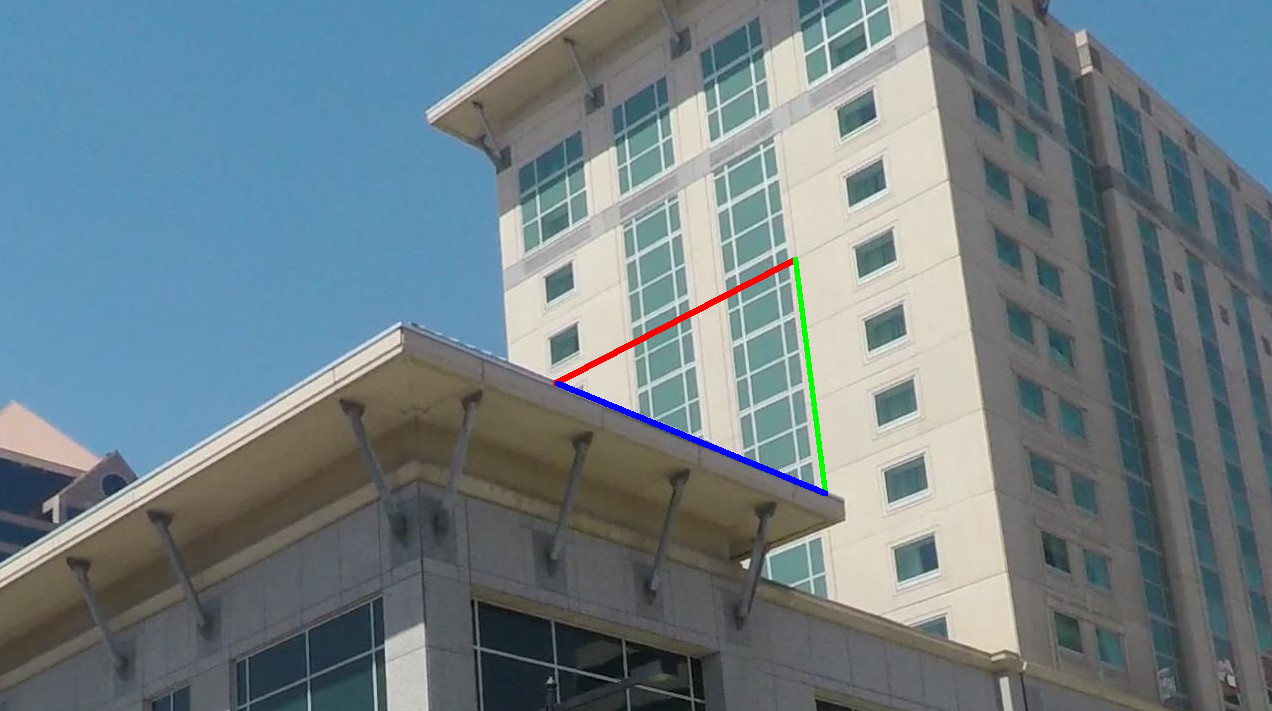}
    \caption{A 3-cycle. The line colors red, green, blue correspond to the $x$, $y$, $z$ directions. Such a set of intersections is not physically realizable.}
    \label{fig:3-cycle}
\end{figure}

%\subsection{Cycle Constraints}
In Fig.~\ref{fig:3-cycle}, we show a real image with cycle graph of length three, with an $x$-line, which intersects with a $y$-line, which intersects with a $z$-line, which, in turn intersects with the $x$-line -- an $xyz$ 3-cycle.
At most 2 of the intersections in the cycle can be true intersections, because it is not possible to start at a point, travel first along the $x$ axis, then along $y$, and finally along $z$, and end up at the same location.

%Attempting to solve the LP naively would not make sense, because such a line graph can not correspond to a physically realizable set of lines, let alone the actual lines of the scene.

If the lines in the example above were, for instance, $l_i,l_j,l_k$, the constraint can be expressed as
\begin{equation} \label{eqn:3cyc-con}
    b_{ij} + b_{jk} + b_{ik} \leq 2.
\end{equation}

\noindent
{\bf 3-Cycles.}
The only type of 3-cycle that can exist in a line graph is the one described in the example above, the $xyz$-type, because intersections between lines with the same label are not considered in the line graph.
For each such 3-cycle, we add a constraint as described by equation~\ref{eqn:3cyc-con}.

\noindent
{\bf 4-Cycles.}
With cycles of length 4, we have two types: the $xyxy$-type (or cyclic permutations $yzyz$, $xzxz$), and the $xyzy$-type (or its permutations).
Now in an $xyxy$ type cycle, it \emph{is} possible for all of the intersections to be valid, and hence we do not add any constraints for this type of cycle.

However, in an $xyzy$-type cycle, at most 3 of the intersections can be real, and we add the constraint
\begin{equation}\label{eqn:4cyc-con}
    b_{ij} + b_{jk} + b_{kl} + b_{li} \leq 3, 
\end{equation}
where $l_i,l_j,l_k,l_l$ are the lines involved, in order.

\subsection{Planarity Constraints}

In addition to the cycle constraints, we implement another type of constraints, which we call the planarity constraints.
Consider a pair of parallel $x$-lines. They can either (a) form an $xy$-plane, or (b) form an $xz$-plane, or (c) not form a Manhattan plane. If they form an $xy$-plane, then any $y$-line must intersect with either both $x$-lines, or none of them (as long as this $y$ line shares an edge with both of our $x$-lines in the line graph). Also, any $z$ can intersect with at most one of the $x$-lines.

We formulate this constraint as follows. For each such pair $\{l_k,l_l\}$ with direction $\mathcal{D}_k = \mathcal{D}_l = d$, we define two Boolean variables, $\pi_{kl}^{(e)}$ and $\pi_{kl}^{(f)}$, corresponding to the other two directions $\{x,y,z\} \setminus \{\mathcal{D}_k\}$.
A value of $1$ for these variables represents that the plane is parallel to the corresponding direction. For instance, for a pair of $y$-lines $l_k$ and $l_l$, the variables $\pi_{kl}^{(x)}$ and $\pi_{kl}^{(z)}$, correspond to the $x$ and $z$ directions respectively.
A value of 1 for $\pi_{kl}^{(x)}$ indicates that the two lines $l_k$ and $l_l$ form an $xy$ plane, and a value of 1 for $\pi_{kl}^{(z)}$ would indicate that the two lines form a $yz$ plane. Of course, it's possible for the lines to not form a Manhattan plane, and so at most one of these variables may take the value 1, i.e.\
\begin{equation}
    \pi_{kl}^{(e)} + \pi_{kl}^{(f)} \leq 1,\quad\{e,f\} = \{x,y,z\} \setminus \{d = \mathcal{D}_k\}. \label{eqn:pla-cons-1}
\end{equation}

Consider a line $l_m$, such that $(k,m) \in \mathcal{E}$, and $(j,m) \in \mathcal{E}$, assuming w.l.o.g.\ that $\mathcal{D}_m = e$, we write
\begin{align}
    \text{if } \pi_{kl}^{(e)} & = 1: \quad b_{km} = b_{lm}, \\
    \text{if } \pi_{kl}^{(f)} & = 1: \quad b_{km} + b_{lm} \leq 1.
\end{align}

Since $\pi_{kl}^{(e)}, \pi_{kl}^{(f)}$ are themselves variables, we rewrite the constraints as
\begin{align}
    \pi_{kl}^{(e)} + b_{km} & \leq 1 + b_{lm}, \label{eqn:pla-cons-2}\\
    \pi_{kl}^{(e)} + b_{lm} & \leq 1 + b_{km} \text{, and} \label{eqn:pla-cons-3}\\
    \pi_{kl}^{(f)} + b_{km} + b_{lm} & \leq 2. \label{eqn:pla-cons-4}
\end{align}

And since we want to have as many Manhattan planes as possible, we define another objective term
\begin{equation}
    \max \sum_{(k,l), \mathcal{D}_k = \mathcal{D}_l} \pi_{kl}^{(e)} + \pi_{kl}^{(f)}. \label{eqn:pla-obj}
\end{equation}

\subsection{Boundary Line Constraints}

\begin{figure}
    \centering
    \includegraphics[width=0.8\columnwidth]{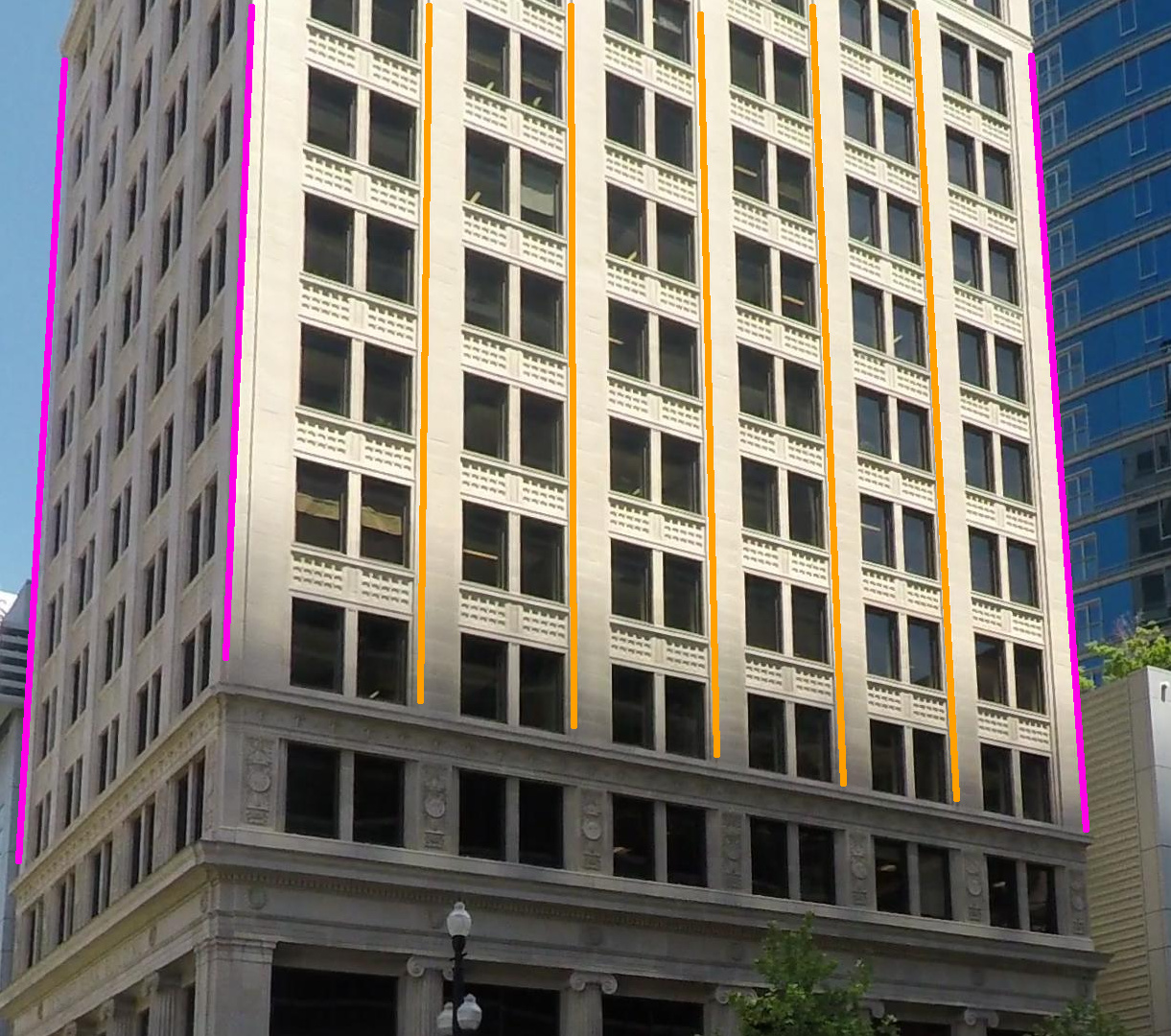}
    \caption{Boundary (magenta) and non-boundary (orange) lines are shown on a typical outdoor scene.}
    \label{fig:boundary}
\end{figure}

Since we are primarily concerned with piece-wise planar scenes, we design a constraint that explicitly enforces this.
Each line in the scene is either at an intersection of two planes in the scene (a \emph{boundary} line), or it is in the middle of a plane (a \emph{non-boundary} line), as shown in figure~\ref{fig:boundary}.

If a line is a non-boundary line, it can intersect with lines from only a particular direction, i.e.\ if a $y$-line is in the middle of a $yz$-plane, then it is not allowed to intersect with any $x$-line.

For each line $l_i$, we define the Boolean variable $B_{i}$, where a value of $1$ indicates that $l_i$ is on the boundary. For all pairs of lines $\{l_m,l_n\}$ such that $(i,m) \in \mathcal{E}$, $(i,n) \in \mathcal{E}$, and $\mathcal{D}_m \neq \mathcal{D}_n$, we enforce the constraint
\begin{equation}
    b_{im} + b_{in} \leq 1 + B_i. \label{eqn:bou-cons}
\end{equation}
That is, if $l_i$ is a boundary line ($B_i = 1$), we would allow it to intersect with both lines, but if it's a non-boundary line, it can intersect with at most one of those lines.

Finally, in accordance with the piece-wise planar assumption, we would like as few lines to be boundary lines as possible, and we define yet another objective term
\begin{equation}
    \max \sum_{i \in \mathcal{V}} (1 - B_i). \label{eqn:bou-obj}
\end{equation}

\subsection{MILP Formulation}

Since all the constraints and the objective terms are linear, this is a mixed integer linear program (MILP).
The full MILP can be written as
\begin{align}
    & \max \Bigg\{ \sum_{(i,j)} w_{ij} b_{ij} + \mu_1 \left(\sum_{\mathcal{D}_k = \mathcal{D}_l} \pi_{kl}^{(e)} + \pi_{kl}^{(f)}\right) \nonumber \\
    & \qquad + \mu_2 \sum_{i} (1 - B_i) \Bigg\}, \label{eqn:MILP-obj} \\
    & \text{subject to} \nonumber \\
    & \left\lvert \lambda_{ia} d_{ia\alpha} - \lambda_{jb} d_{jb\alpha} \right\rvert \leq L(1-b_{ij}), \quad a,b \in \{1,2\}, \nonumber \\
    & \qquad \alpha = \{x,y,z\} \setminus \{\mathcal{D}_i, \mathcal{D}_j\}, \ (i,j) \in \mathcal{E} \tag{\ref{eqn:MILP-cons-1}}\\
    & \left\lvert \lambda_{i1} d_{i1\beta} - \lambda_{i2} d_{i2\beta} \right\rvert \leq 0, \nonumber \\
    & \qquad \beta \in \{x,y,z\} \setminus \{\mathcal{D}_i\}, i \in \mathcal{V} \tag{\ref{eqn:MILP-cons-2}}\\
    & \lambda_{ia} \geq 1,  \quad a \in \{1,2\}, i \in \mathcal{V}, \tag{\ref{eqn:MILP-cons-3}} \\
    & b_{ij} + b_{jk} + b_{ik} \leq 2, \quad \{ (i,j),(j,k),(k,i) \} \subseteq \mathcal{E} \tag{\ref{eqn:3cyc-con}} \\
    & b_{ij} + b_{jk} + b_{kl} + b_{il} \leq 3, \nonumber \\
    & \qquad \{ (i,j),(j,k),(k,l),(i,l) \} \subseteq \mathcal{E} \tag{\ref{eqn:4cyc-con}} \\
    & \pi_{kl}^{(e)} + \pi_{kl}^{(f)} \leq 1,\quad\{e,f\} = \{x,y,z\} \setminus \{d = \mathcal{D}_k\}. \tag{\ref{eqn:pla-cons-1}} \\
    & \pi_{kl}^{(e)} + b_{km} \leq 1 + b_{lm}, \tag{\ref{eqn:pla-cons-2}} \\
    & \pi_{kl}^{(e)} + b_{lm} \leq 1 + b_{km}, \tag{\ref{eqn:pla-cons-3}} \\
    & \pi_{kl}^{(f)} + b_{km} + b_{lm} \leq 2, \tag{\ref{eqn:pla-cons-4}} \\
    & b_{im} + b_{in} \leq 1 + B_i, \tag{\ref{eqn:bou-cons}}
\end{align}
where $\mu_1$ and $\mu_2$ are parameters that we tune.

Because of noise in image acquisition and calibration, we may not be able to satisfy the constraints \ref{eqn:MILP-cons-1} and \ref{eqn:MILP-cons-2} exactly, and so we relax these using slack variables.

\subsection{Minimum Spanning Tree}

A sufficient condition to reconstruct all $N$ lines of a line graph is that the graph is a tree, i.e.\ it (a) is connected, and (b) has $N - 1$ edges.
In our approach, we might have more than $N-1$ edges, and we simultaneously identify whether each edge is real ($b_{ij}$), and solve for the $\lambda_{ia}$ variables.
However, since only the edges from the minimal spanning tree are ``necessary'', in terms of evaluating the approach we look at the fraction of real intersections in the MST.
% Even after applying the loop and planarity constraints, where we either select ($w_{ij} = 1$) or deselect ($w_{ij} = 0$) an edge, we might have more than $N-1$ edges. It is possible that some of these intersections are not realizable. Thus, after we solve the function~\ref{eqn:LP-obj}, not all of the $s_{ij}$ variables will have value 0. To deal with this issue, we take a two pass approach. After the first pass of the MILP, we compute the minimum spanning tree (MST) of the line graph, with $s_{ij}$ as the edge weights.
% This is like saying that the intersections with the lowest $s_{ij}$ values are the ones identified by the MILP as being true intersections. We then do a second pass of the LP, this time using only intersections from the MST.
The full procedure is outlined in Algorithm~\ref{alg:full-alg}.

\begin{algorithm}[t]
    \begin{algorithmic}[1]
        \INPUT Image $I$
        \OUTPUT Set of lines $\mathcal{L}$ in 3D
        \State $\mathcal{L}_{2D0} = \{(p_{i1}, p_{i2}) \} \gets \text{DetectLineSegments}(I)$
        \State $(V_x, V_y, V_z) \gets \text{EstimateVanishingPoints}(\mathcal{L}_{2D})$
        \State $\mathcal{D} = \{\mathcal{D}_{i} : i \in \mathcal{L}_{2D} \} \gets \text{Label}(\mathcal{L}_{2D0}, V_x, V_y, V_z)$
        \State $\mathcal{G}_0 = (\mathcal{L}_{2D0}, \mathcal{E}_0) \gets \text{CreateLineGraph}(\mathcal{L}_{2D0}, \mathcal{D})$
        \State $\mathcal{G} = (\mathcal{L}_{2D}, \mathcal{E}) \gets \text{LargestConnectedComponent}(\mathcal{G}_0)$
        \State $b_{ij}, \lambda_{ia} \gets \text{SolveMILP}(\mathcal{L}_{2D}, \mathcal{E})$ \hfill (Equation~\ref{eqn:MILP-obj})
        \State $P_{ia} \gets \lambda_{ia} R^{-i} K^{-1} p_{ia}$
        \State $\mathcal{L} \gets \{ (P_{i1}, P_{i2})\}$
        % \State $\mathcal{G}_M = (\mathcal{L}_{2D}, \mathcal{E}_M) \gets \text{KruskalMST}(\mathcal{G}, s_{ij})$
    \end{algorithmic}
    \caption{Line Reconstruction Algorithm}
    \label{alg:full-alg}
\end{algorithm}

\section{Experiments}
% !TEX root = ../paper.tex
%
\subsection{Data Collection}

One of the main contributions of this work is the dataset -- 50 high resolution (4K) distortion-corrected photos of urban scenes acquired using a calibrated GoPro, with detected lines. Note that this dataset is far more challenging than York Urban database~\cite{denis08} used in ~\cite{Ramalingam2013_lifting}, since the images are captured from a car-mounted camera with a large variation in the depth of the buildings in the images. The intersections of the lines have been (manually) marked as real or fake.
To calibrate the GoPro and correct the images for distortion, we use MATLAB's Camera Calibration toolbox. For line segment detection we used code from \cite{MCMLSD}. Finally, we labeled the line intersections as real or fake, using a simple UI we built (in MATLAB).

\subsection{Implementation}

Our line reconstruction code is written in C++, using Eigen~\cite{Eigen} for the geometry part, and Boost Graph~\cite{BoostGraph} for the graph functionality.

To solve the MILP, we use Gurobi Optimizer~\cite{Gurobi}, which uses a branch and bound algorithm, which we terminate early if necessary based on a time budget of 300 seconds. In most of our examples, however, we observe that the optimization runs to completion far quicker, with an average reconstruction time of 18 seconds per image on a desktop computer, and only 3 of the 50 requiring early termination. This implies that we achieve the globally optimal solution for the objective function in most of the images.

For constructing the line graph, we extend all lines by 30 pixels on both sides, and for the parameters corresponding to the planarity and boundary constraints, we use the values $\mu_1 = 0.5$ and $\mu_2 = 10$ respectively, tuned using a simple grid search.

In practice, the reconstruction is relatively insensitive to parameter tuning, with the percentage accuracies decreasing only by at most $0.4\%$ even for a worse choice of parameters (as long as they are non-zero).

\section{Results}
% !TEX root = ../paper.tex
%
We evaluate our method both quantitatively and qualitatively. For the former, we report the fraction of intersections in the minimum spanning tree that have a ground truth label of ``real'', as opposed to intersections that have a value of fake. We report these numbers with and without all of the constraints we have described previously. These results are summarized in table~\ref{tab:results}. If $n_a$ of the $N_a$ intersections in the MST from image $a$ are real, then the mean accuracy is defined as $\frac{\sum_a n_a}{\sum_a{N_a}}$, and the norm. accuracy is defined as $\mean_a(\frac{n_a}{N_a})$. The ablation study demonstrates that each set of constraints provide some benefits over the baseline~\cite{Ramalingam2013_lifting}.
We also report some statistics of our dataset in table~\ref{tab:stats}.

\begin{table}[t]
    \centering
    \caption{Ablation study for the various constraints over the baseline method.}
    \label{tab:results}
    \begin{tabular}{c c c}
        \hline
        Method & Mean Acc. \% & Norm. Acc. \% \\
        \hline\hline
        Baseline~\cite{Ramalingam2013_lifting} & 80.5815 & 78.3001 \\
        Boundary (B) & 82.8908 & 80.5984 \\
        Cycles (C) & 81.4417 & 79.2585 \\
        Planarity (P) & 80.9835 & 78.6172 \\
        B + C & 82.9095 & 80.5094 \\
        B + P & 83.003 & 80.5325 \\
        C + P & 81.5819 & 79.4258 \\
        \textbf{B + C + P} & \textbf{83.2087} & \textbf{80.6073} \\
        \hline
    \end{tabular}
\end{table}

\begin{table}[t]
    \centering
    \caption{Statistics of lines, line graphs, and true intersections.}
    \label{tab:stats}
    \begin{tabular}{| p{5cm} | c |}
        \hline
        Total \# labeled images & 50 \\ \hline
        Avg \# line segments in largest connected component & 214.9 \\ \hline
        Avg \# of connections in line graph & 867.22 \\ \hline
        Avg \# correct connections & 751.78 \\ \hline
        Avg \# incorrect connections & 115.44 \\ \hline
    \end{tabular}
\end{table}

Selected images and reconstructed 3D models are presented in table~\ref{tab:resultImages}. The first five rows are some of the good results, and the last two rows are representative of the failure cases. Note that the reconstruction only shows the largest connected component in the line graph. %In particular, as mentioned earlier, disconnected components can not be handled with a single-view approach, and if there are some lines in the image connecting such components, the reconstruction can have unexpected artifacts.

\begin{table*}[tp]
    \setlength\tabcolsep{0pt}
    \caption{The first five rows are some of the good results, and the last two rows are the failure cases.}
    \label{tab:resultImages}
    \begin{tabular}{cccc}
        Image & Line Segments & 3D View 1 & 3D View 2 \\
        \hline
        \includegraphics[width=.24\textwidth]{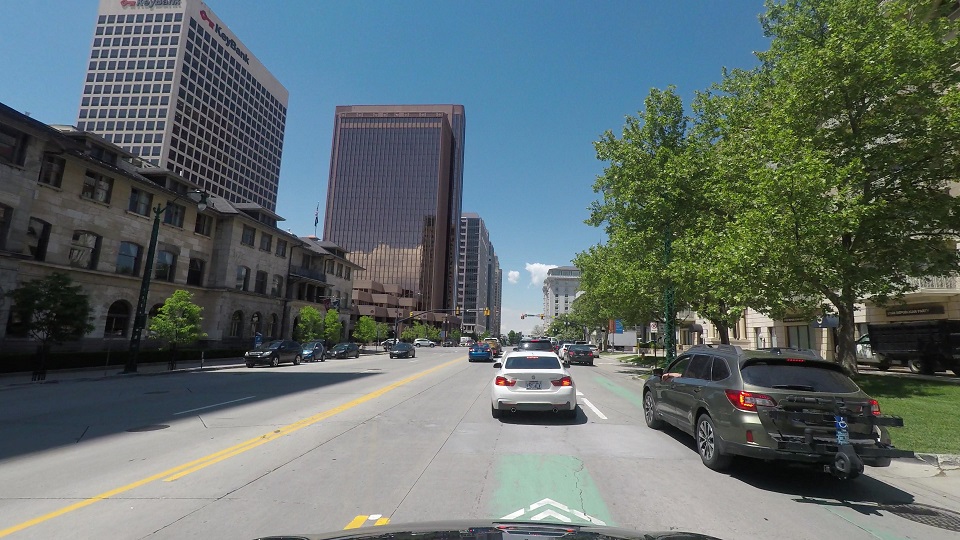} &
        \includegraphics[width=.24\textwidth]{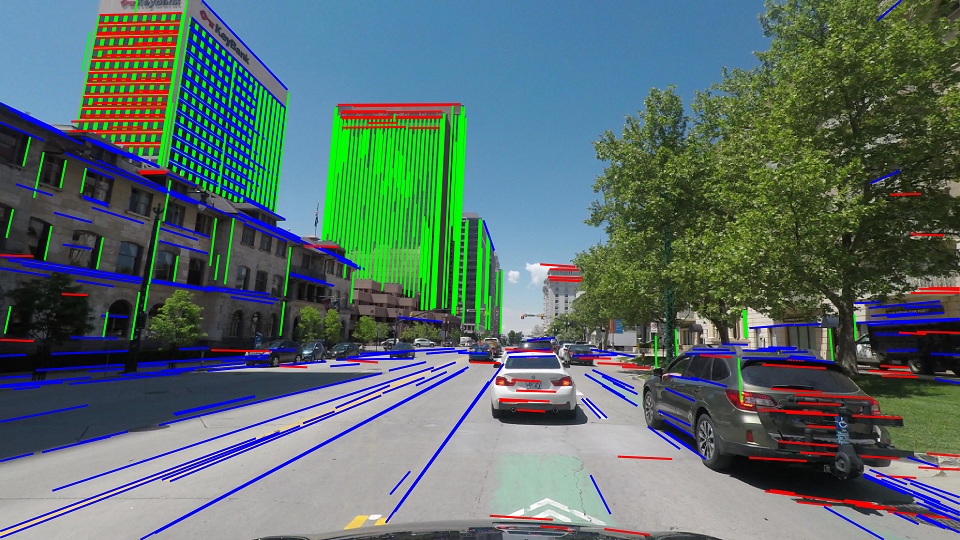} &
        \includegraphics[width=.24\textwidth]{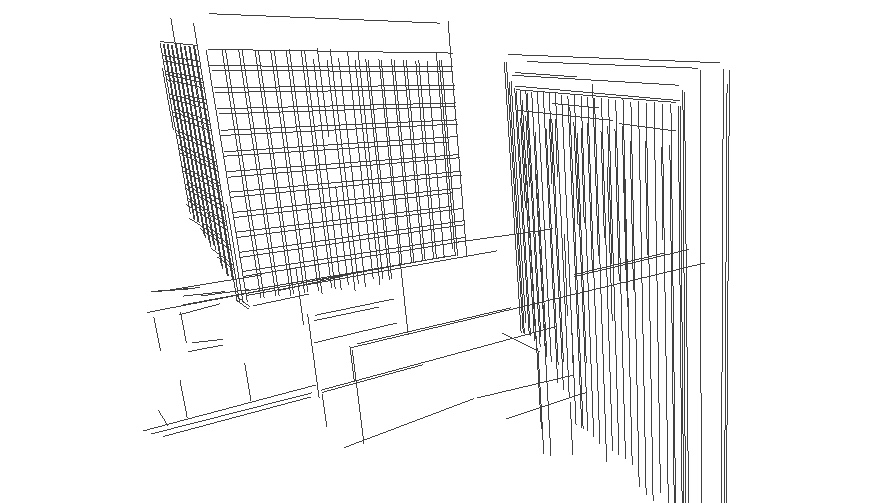} &
        \includegraphics[width=.24\textwidth]{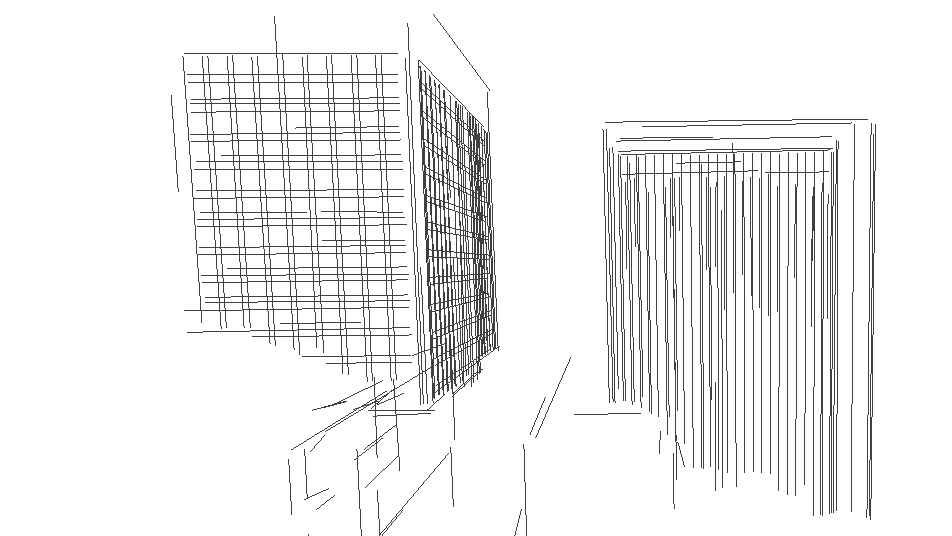} \\

        \includegraphics[width=.24\textwidth]{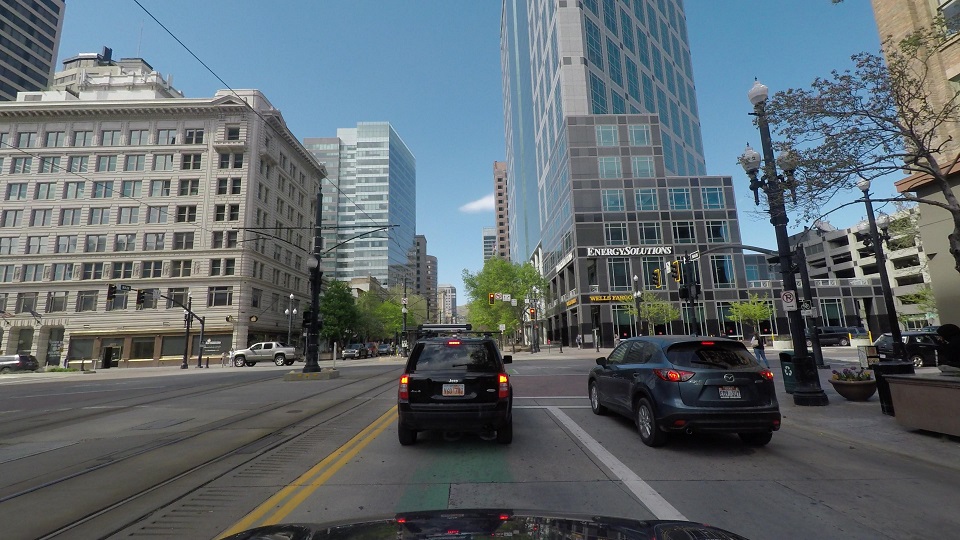} &
        \includegraphics[width=.24\textwidth]{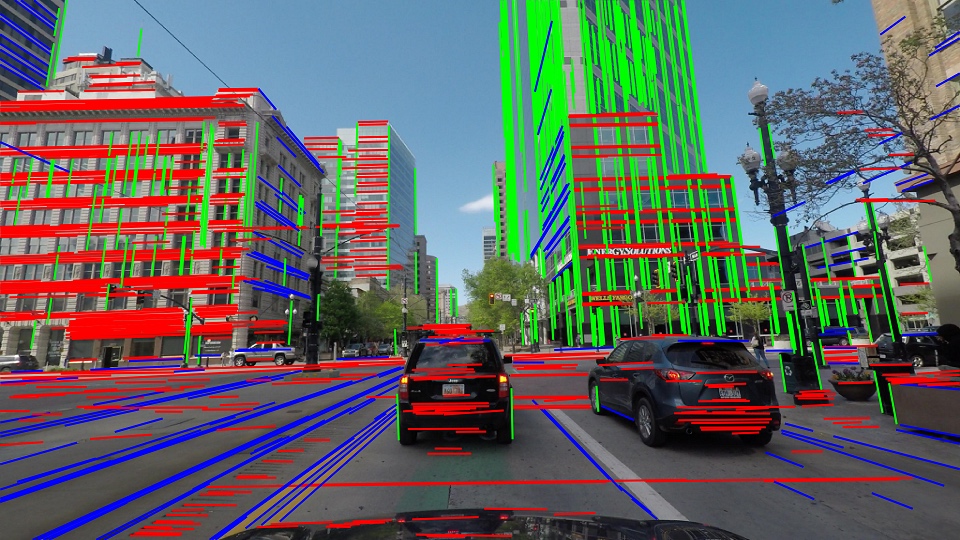} &
        \includegraphics[width=.24\textwidth]{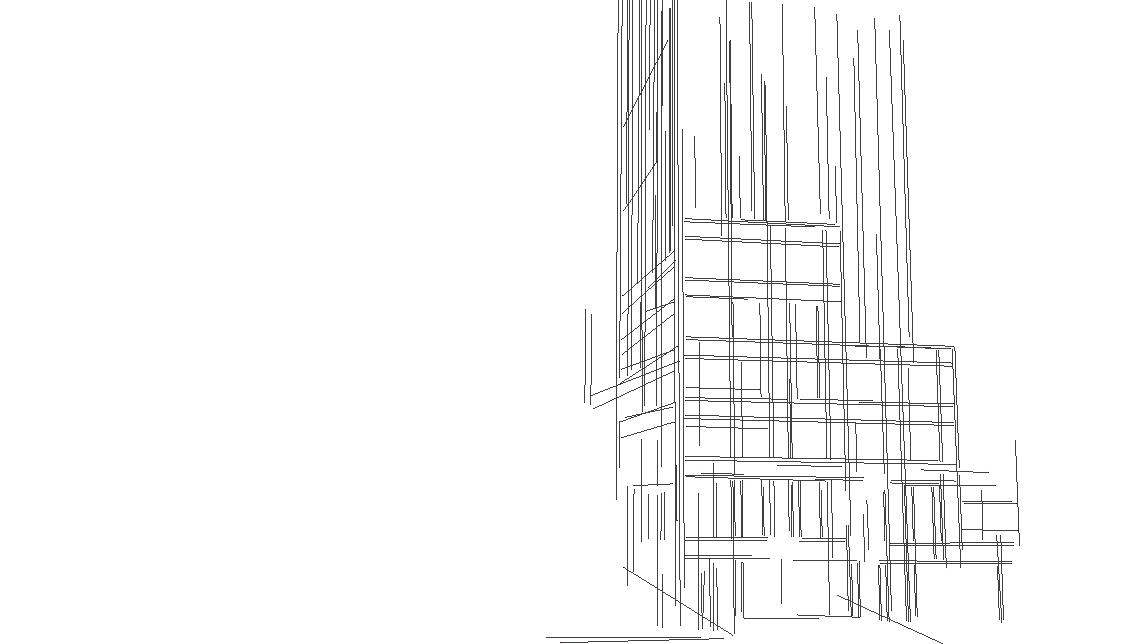} &
        \includegraphics[width=.24\textwidth]{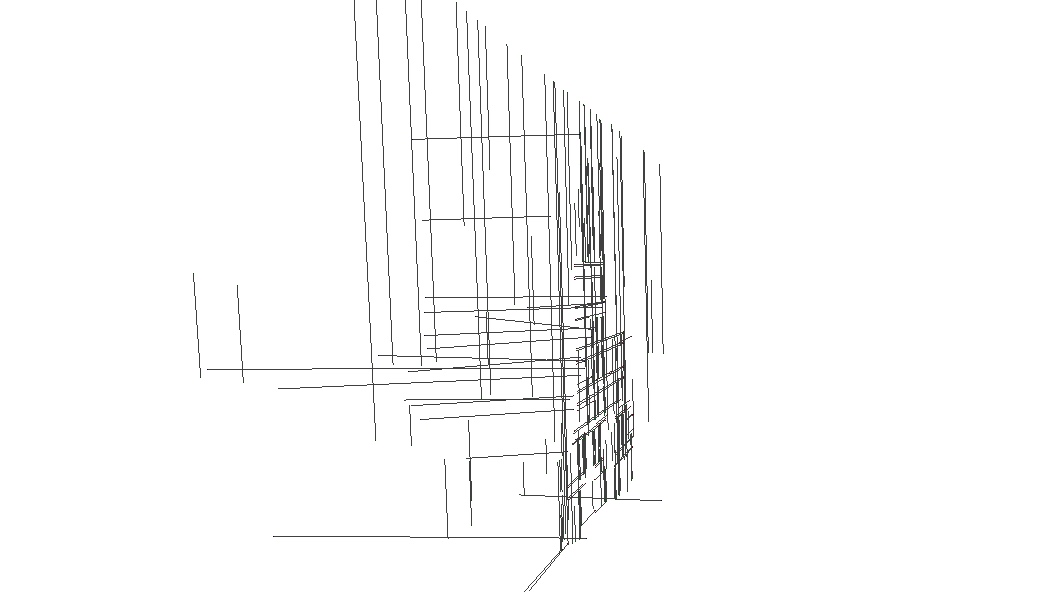} \\

        \includegraphics[width=.24\textwidth]{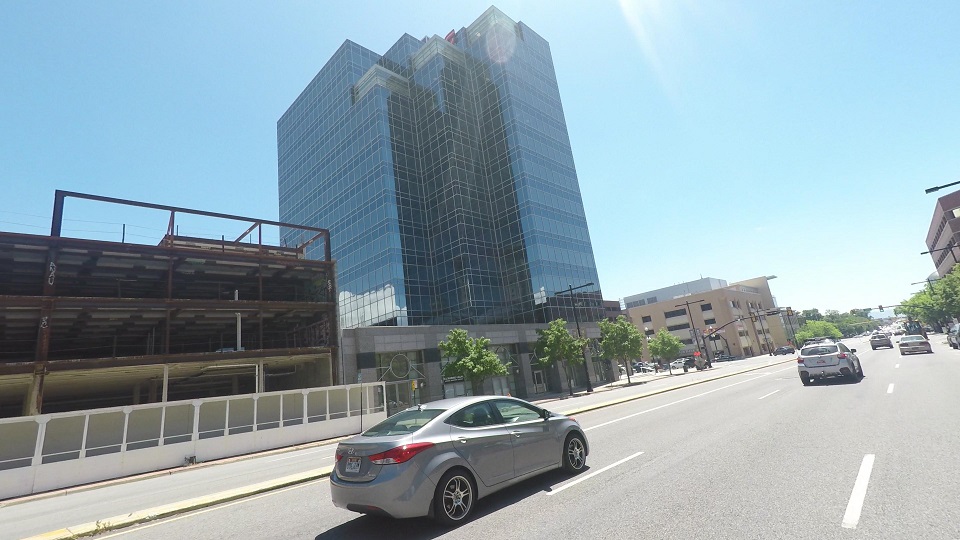} &
        \includegraphics[width=.24\textwidth]{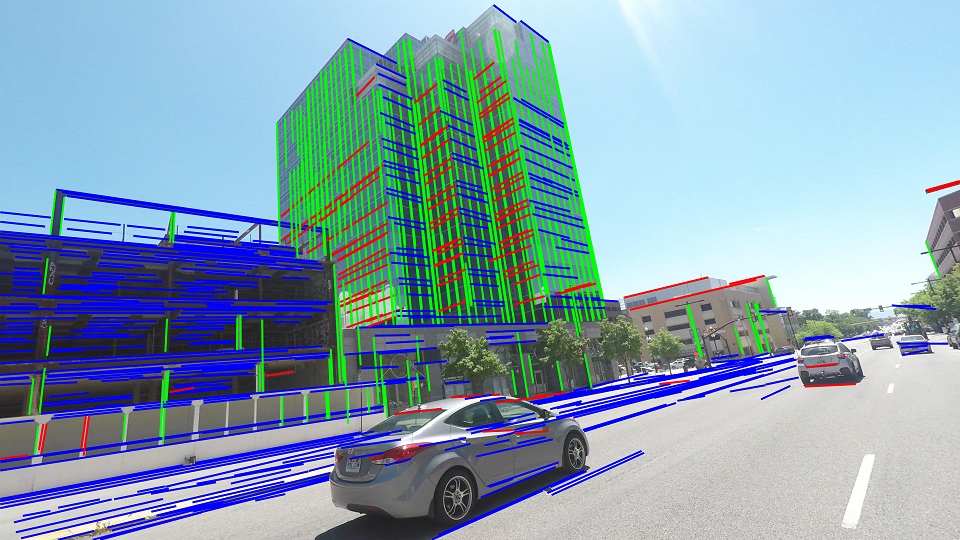} &
        \includegraphics[width=.24\textwidth]{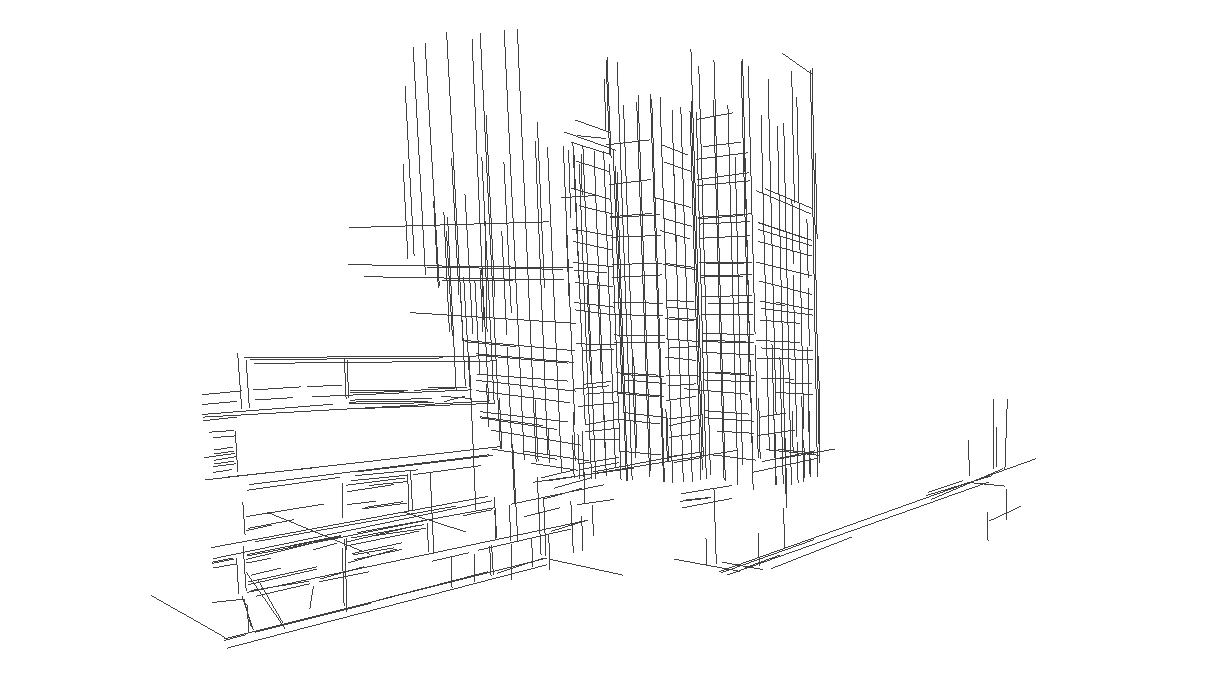} &
        \includegraphics[width=.24\textwidth]{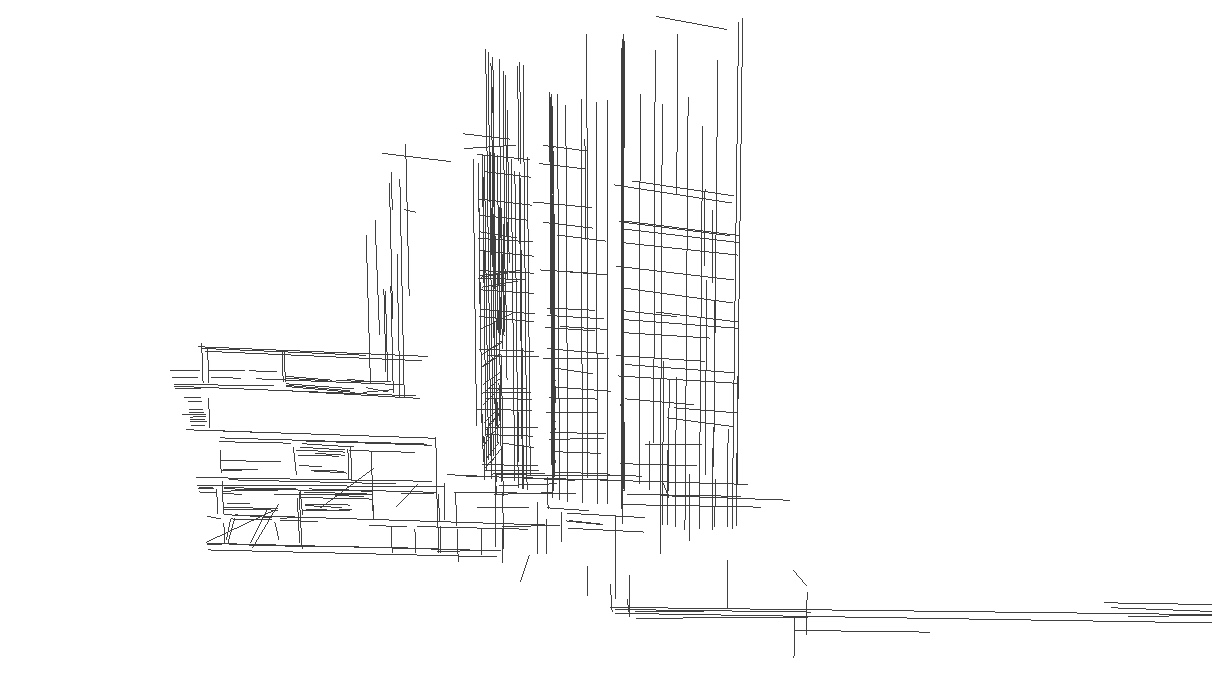} \\

        \includegraphics[width=.24\textwidth]{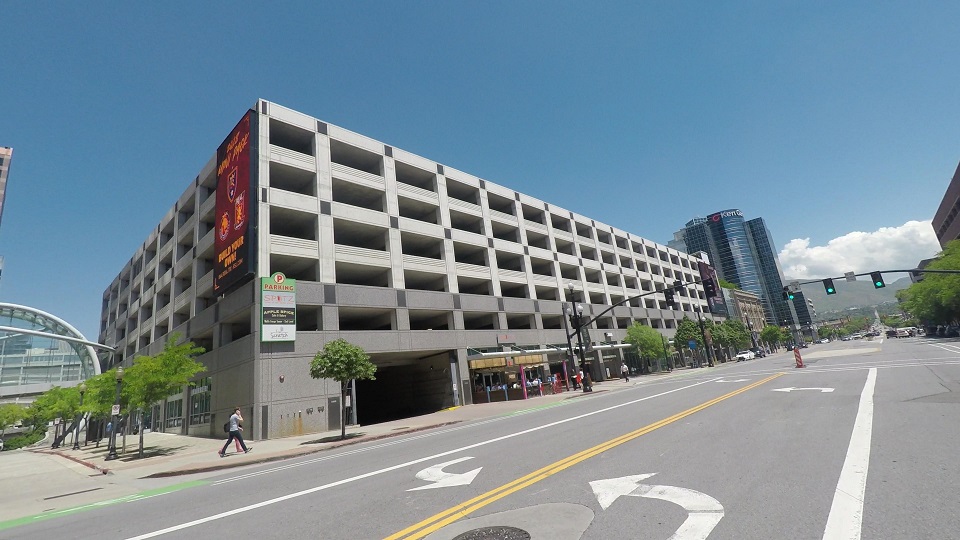} &
        \includegraphics[width=.24\textwidth]{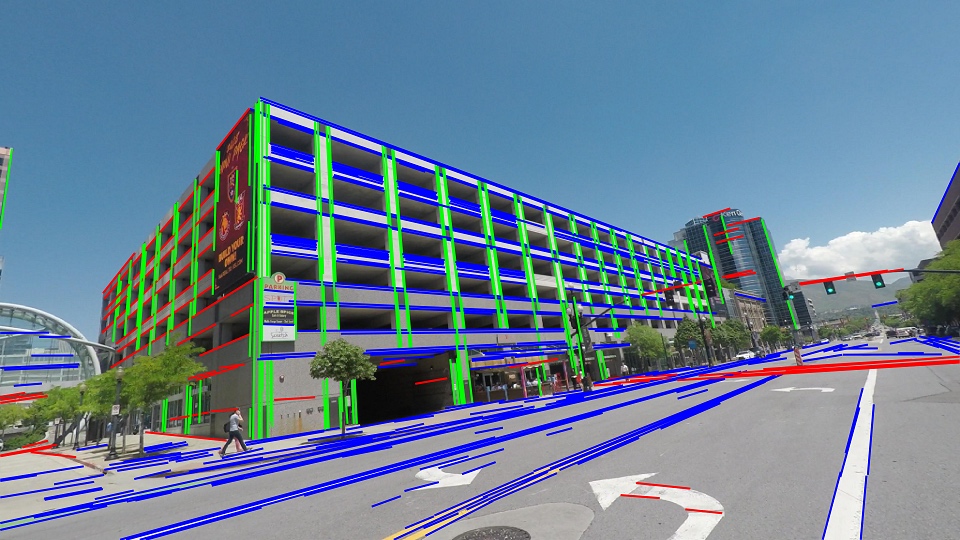} &
        \includegraphics[width=.24\textwidth]{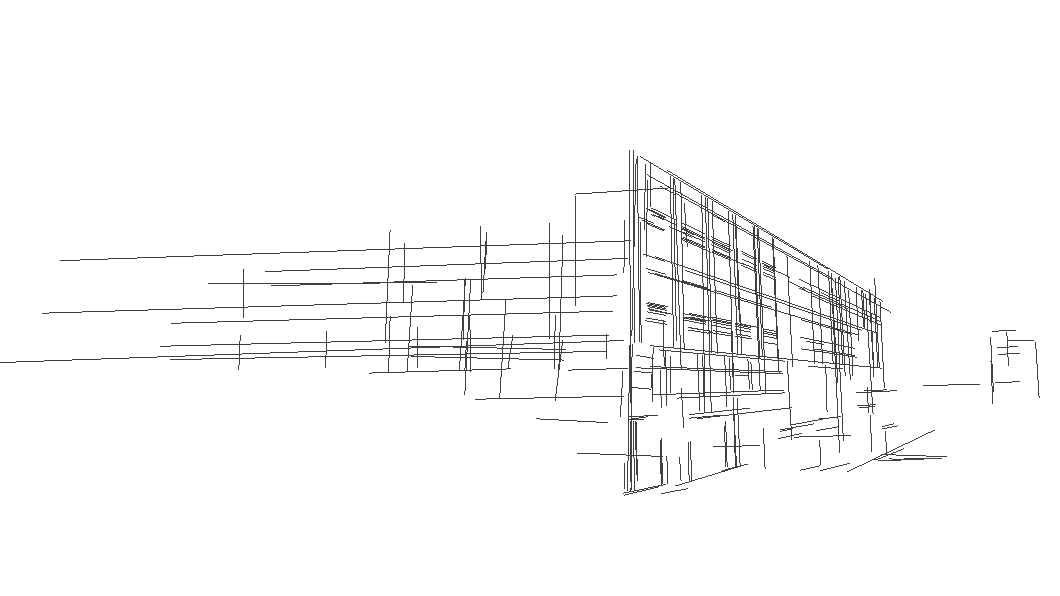} &
        \includegraphics[width=.24\textwidth]{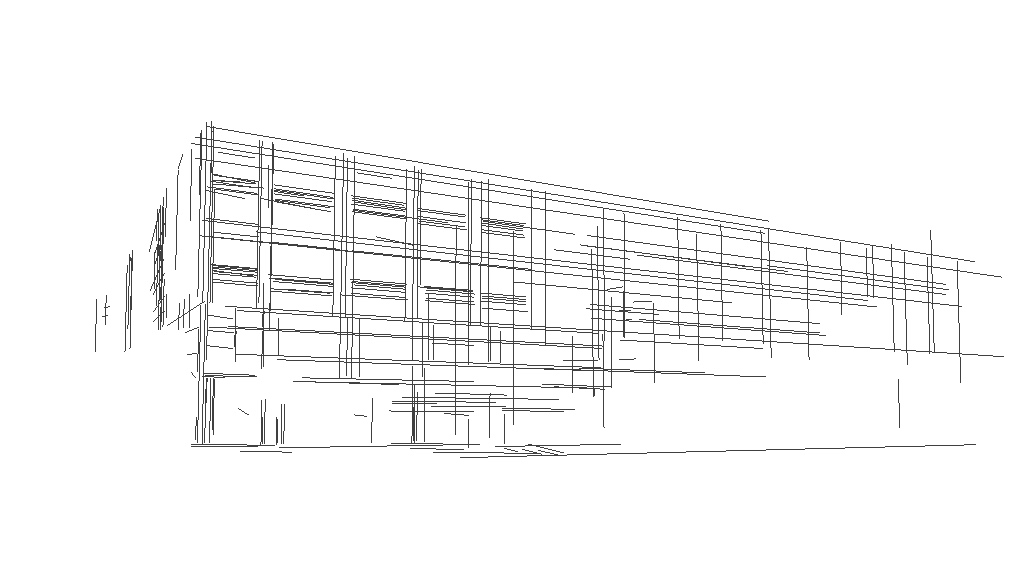} \\

        \includegraphics[width=.24\textwidth]{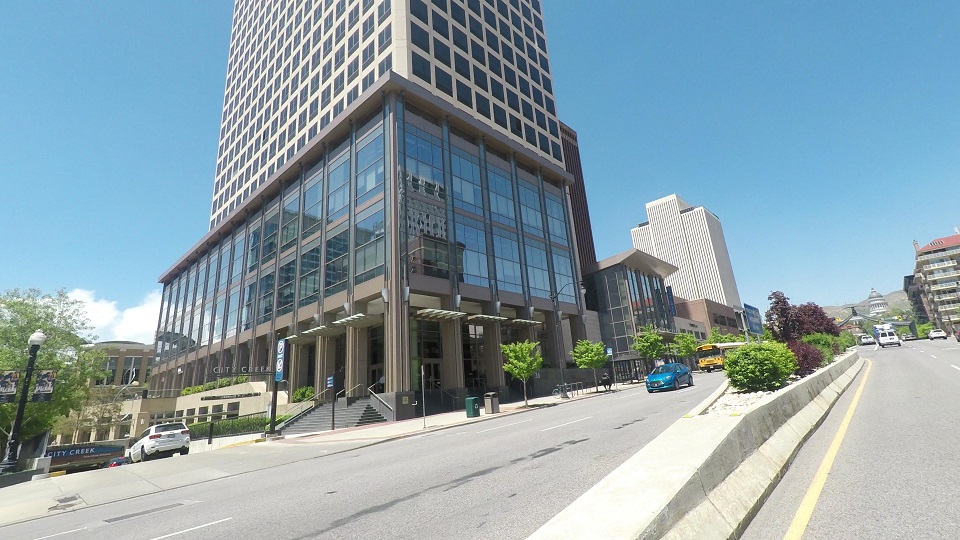} &
        \includegraphics[width=.24\textwidth]{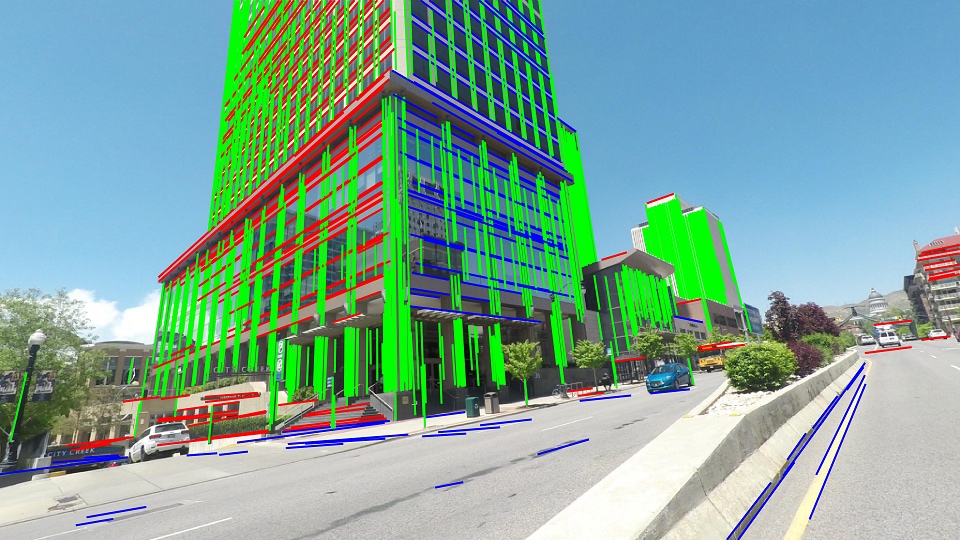} &
        \includegraphics[width=.24\textwidth]{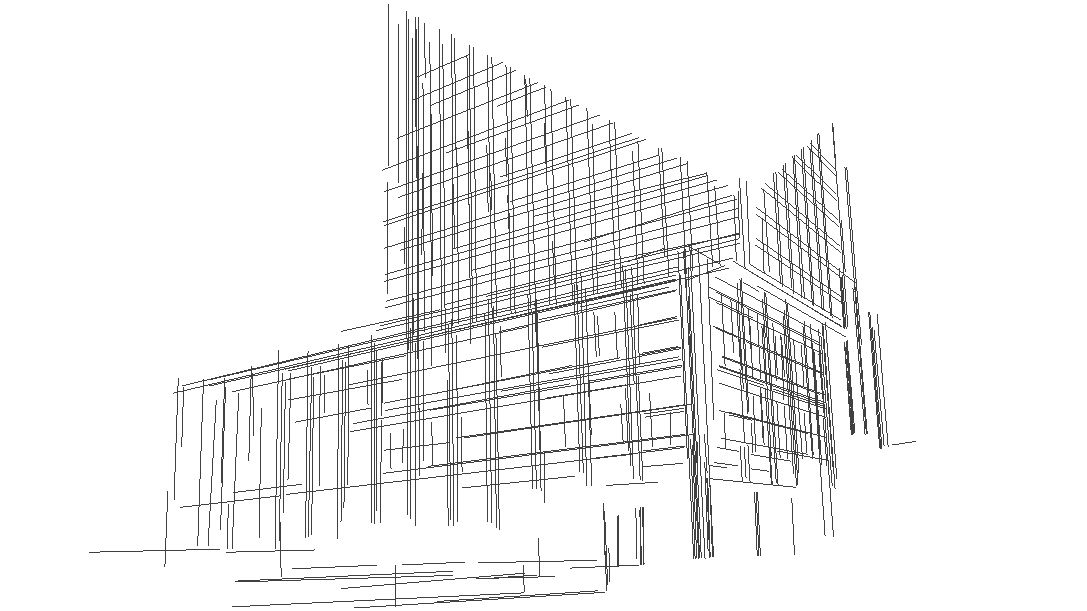} &
        \includegraphics[width=.24\textwidth]{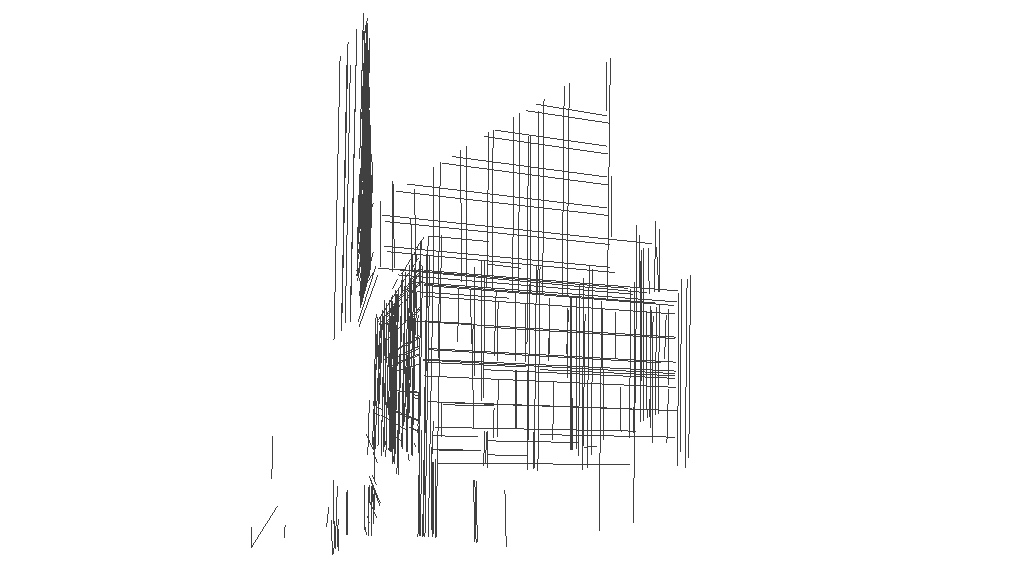} \\

        \includegraphics[width=.24\textwidth]{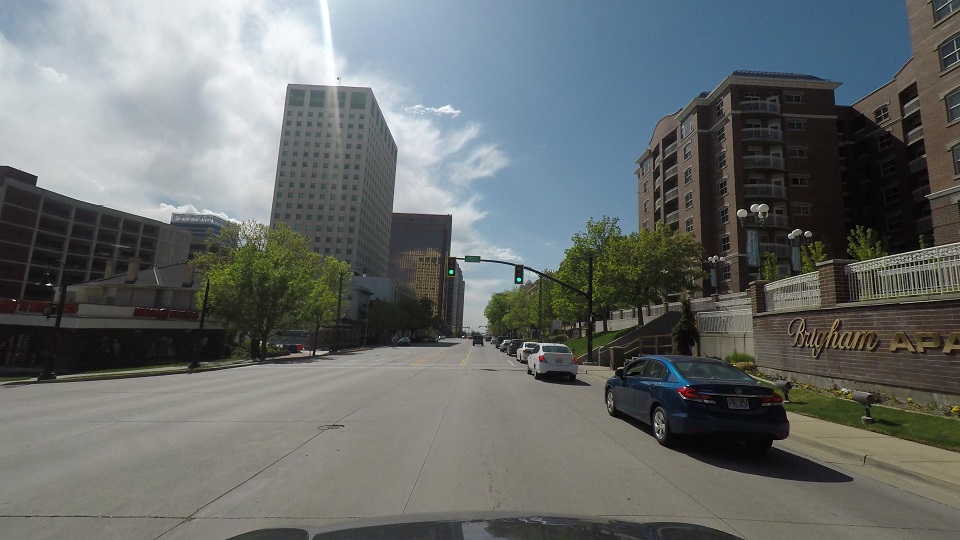} &
        \includegraphics[width=.24\textwidth]{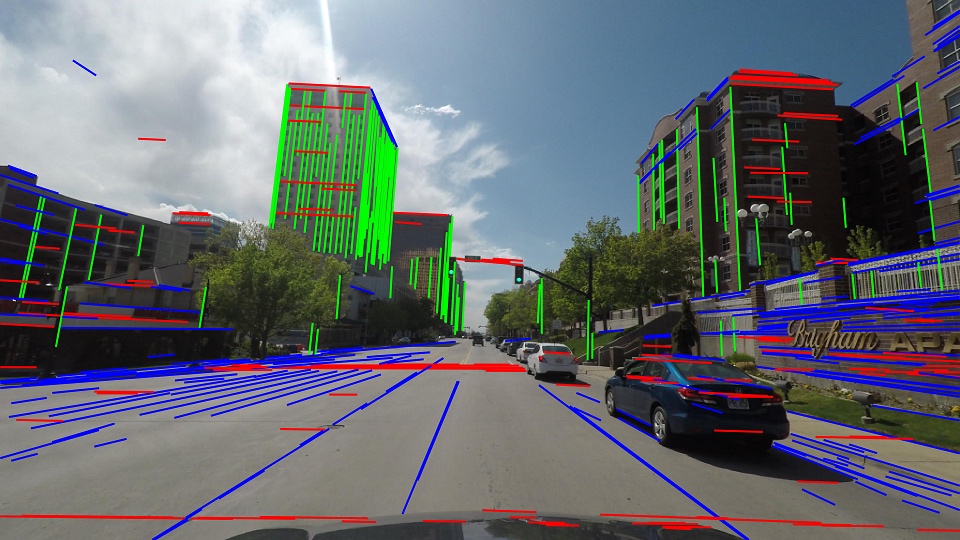} &
        \includegraphics[width=.24\textwidth]{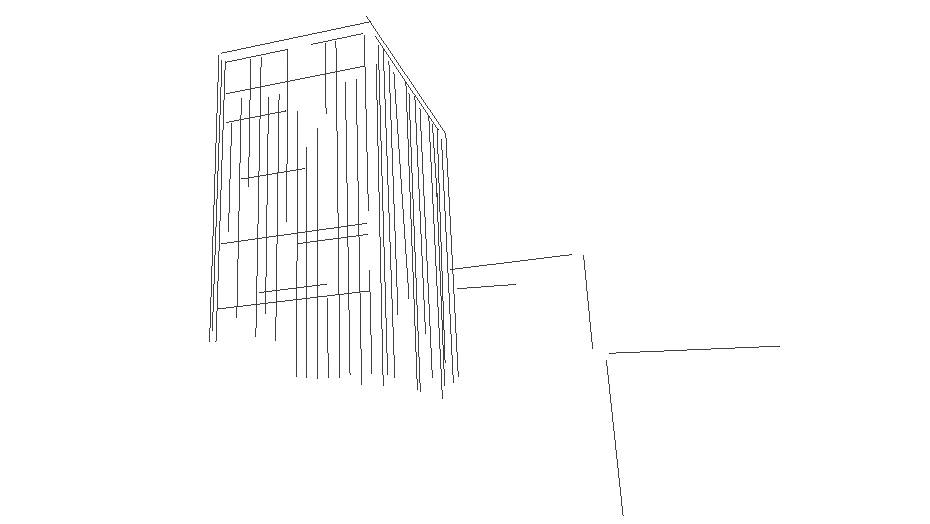} &
        \includegraphics[width=.24\textwidth]{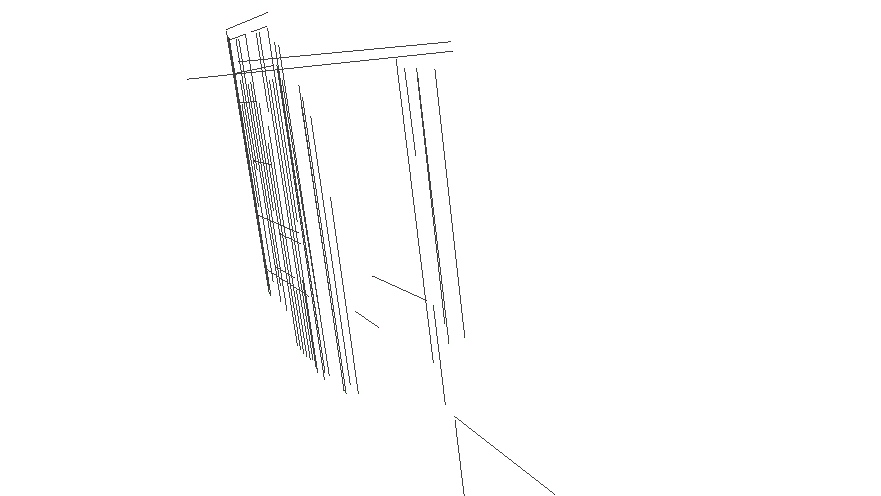} \\

        \includegraphics[width=.24\textwidth]{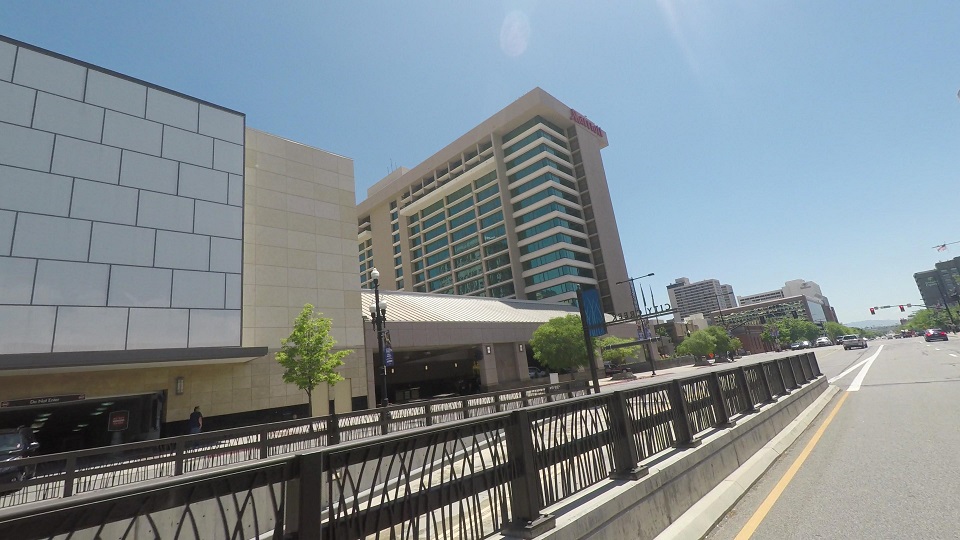} &
        \includegraphics[width=.24\textwidth]{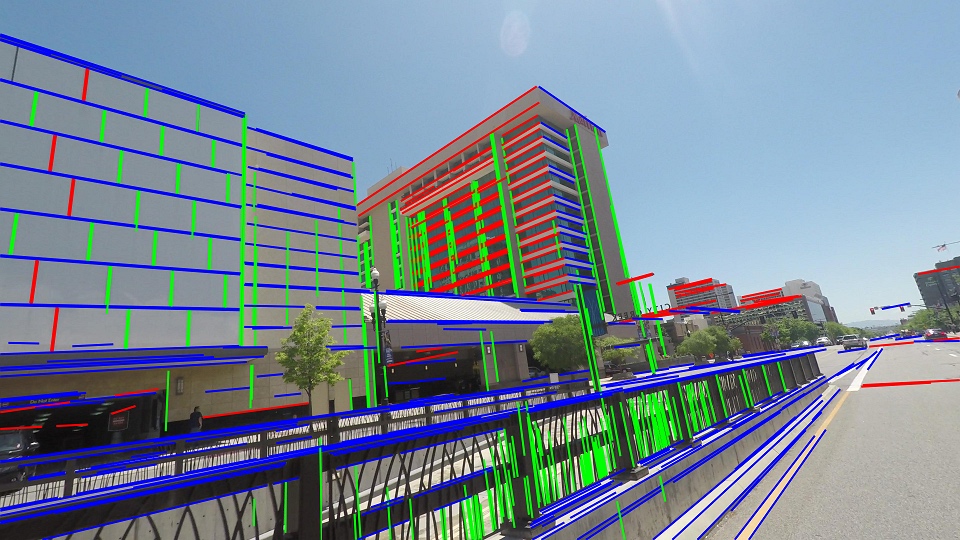} &
        \includegraphics[width=.24\textwidth]{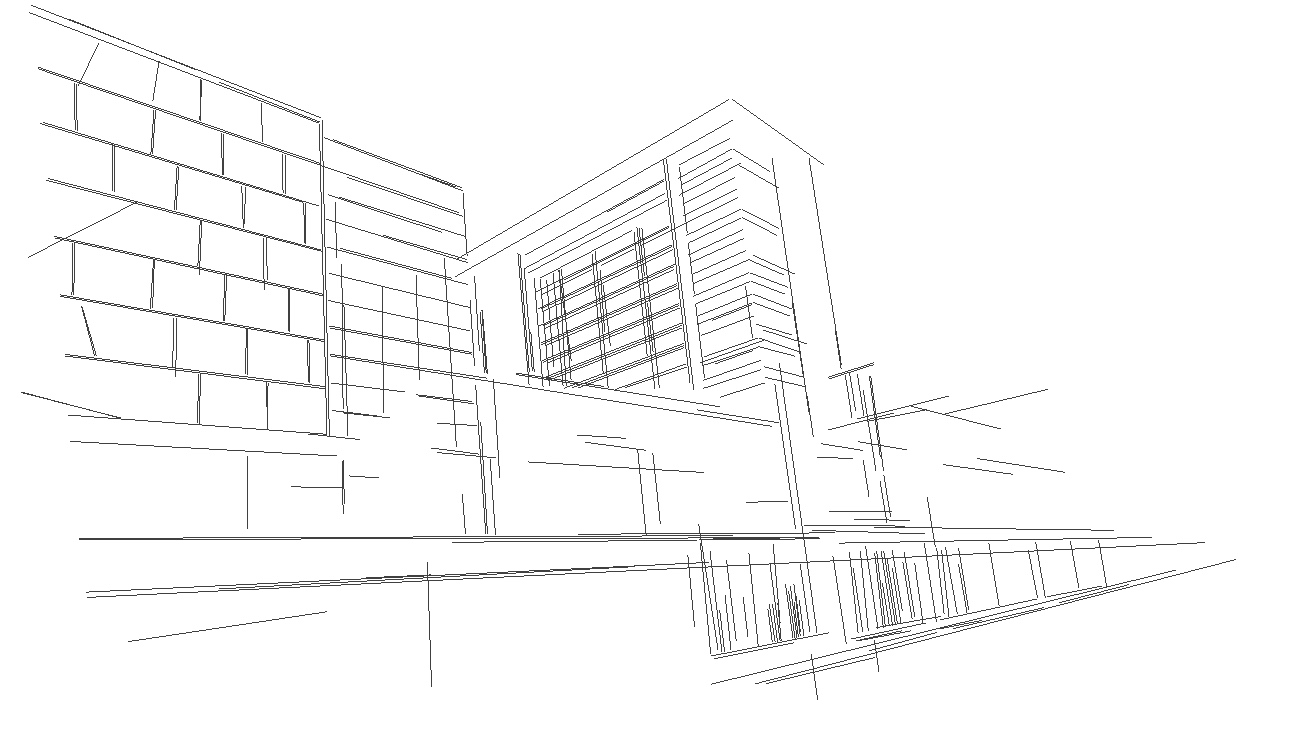} &
        \includegraphics[width=.24\textwidth]{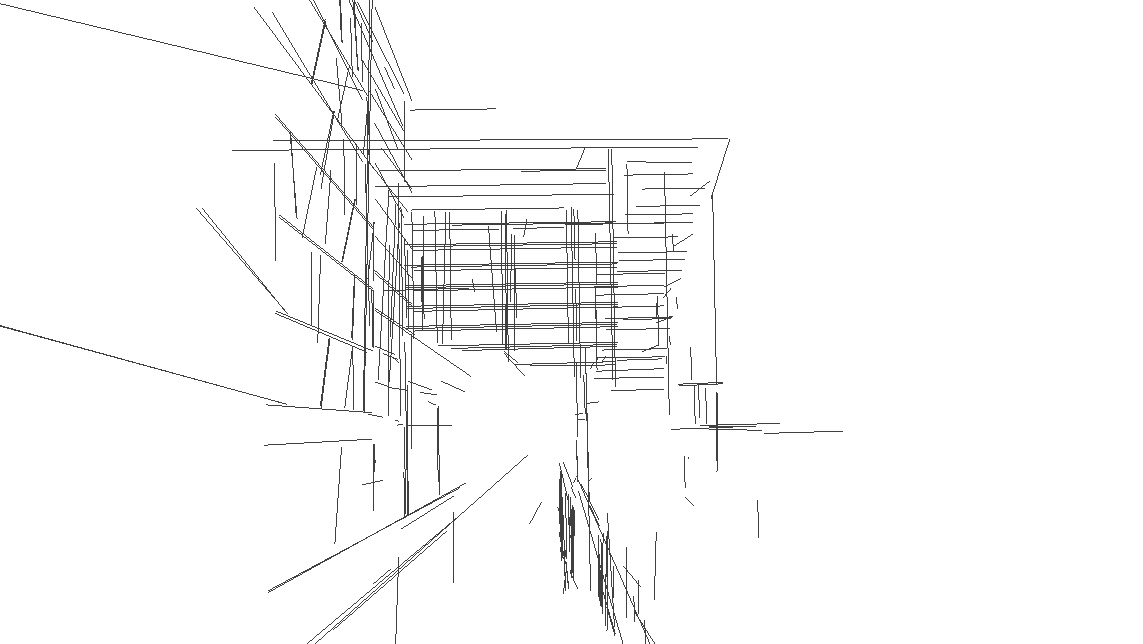} \\
    \end{tabular}
\end{table*}

\section{Discussion}
We show three sets of novel constraints that are applicable for single view line reconstruction. The proposed constraints show improvement over the state-of-the-art techniques. We will release a new challenging dataset for testing line-based reconstruction of Manhattan worlds. We can think of several possible future avenues for this research. First, we plan to incorporate occluding contour lines such as skylines~\cite{ramalingam2010skyline2gps} in the 3D reconstruction pipeline. Since most of the constraints are hard (i.e., they have to be strictly satisfied), the performance was not sensitive to parameter tuning for the constraints. However, it is important to realize that we employ only a fraction of the set of all possible constraints. One possible research would be to identify the subset of most important constraints that enable the most accurate single view 3D reconstruction. Since the approach already recognizes local planes and boundary lines, we can move toward reconstructing a plane-based 3D model. 

\section*{Acknowledgments}
\noindent We thank the reviewers and area chairs for their feedback. 

{\small
\bibliographystyle{ieee}
\bibliography{paper}
}

\end{document}